\documentclass{article}

\PassOptionsToPackage{numbers, compress}{natbib}
\usepackage{multirow}
\usepackage{adjustbox}
\usepackage{booktabs}
\usepackage{xcolor}
\usepackage{array}
\usepackage{hyperref}
\usepackage{tikz}
\usetikzlibrary{positioning, arrows.meta}
\usepackage{makecell}
\usepackage{subcaption}
\usepackage{amsmath}
\usepackage{cleveref}
\usepackage{enumitem}
\usepackage{listings}
\usepackage[normalem]{ulem}
\usepackage{longtable}
\usepackage{wrapfig}
\usepackage{caption}
\usepackage[preprint]{neurips_2026}

\usepackage[utf8]{inputenc} %
\usepackage[T1]{fontenc}    %
\usepackage{hyperref}       %
\usepackage{url}            %
\usepackage{booktabs}       %
\usepackage{amsfonts}       %
\usepackage{nicefrac}       %
\usepackage{microtype}      %
\usepackage{xcolor}         %

\title{Life After Benchmark Saturation: A Case Study of CORE-Bench}

\author{%
  \textbf{Nitya Nadgir}$^{1}$ \quad \textbf{Sayash Kapoor}$^{2}$ \quad \textbf{Kangheng Liu}$^{2}$ \quad \textbf{Peter Kirgis}$^{2}$ \\
  \textbf{Matilda Orona}$^{3}$ \quad \textbf{Stephan Rabanser}$^{2}$ \quad \textbf{Tilman Bayer}$^{1}$ \quad \textbf{Abhishek Shetty}$^{1}$ \quad \textbf{Yue Ling}$^{1}$ \\ 
  \textbf{Derrick Chan-Sew}$^{1}$ \quad
  \textbf{Rumi Nakagawa}$^{1}$ \quad
  \textbf{Saiteja Utpala}$^{2}$ \quad
  \textbf{Zachary S. Siegel}$^{4}$ \quad \\ 
  \textbf{Arvind Narayanan}$^{2}$ \\[4pt]
  \normalfont\small
  $^{1}$Independent \quad $^{2}$Princeton University \quad
  $^{3}$UC Berkeley \quad $^{4}$MIT 
}

\begin{document}
\maketitle

\begin{abstract}
  When a benchmark's accuracy saturates, it is often retired and replaced with a more challenging version. We show that this approach privileges accuracy and misses the opportunity to study six other key dimensions of agent performance: construct validity issues such as shortcuts, out-of-distribution generalizability, efficiency, reliability, the relative importance of the model versus the scaffold, and uplift from human-agent collaboration. We use CORE-Bench Hard, a benchmark for computational reproducibility of scientific code, as a case study to demonstrate that measuring agents along these dimensions yields meaningful insights into agent performance even after accuracy saturates. First, we surface threats to construct validity in CORE-Bench Hard that are difficult to anticipate with less capable agents. We introduce an improved benchmark, CORE-Bench v1.1, and an out-of-distribution task suite, CORE-Bench OOD. Second, we find that despite accuracy saturation, CORE-Bench v1.1 remains useful for measuring efficiency, reliability, model performance, and scaffold performance. Finally, we conduct a small-scale randomized experiment to measure uplift from human-agent collaboration on real-world computational reproducibility tasks. We find a statistically significant speedup by about a factor of two --- likely underestimated due to one-fifth of human-only reproductions reaching the time limit before completing --- and describe various other findings. Together, our contributions present a more rigorous alternative to the dominant accuracy-centric evaluation paradigm. 
\end{abstract}

\section{Introduction}

AI agents are increasingly deployed across a wide range of domains, including customer service~\citep{yao2025taubench}, software engineering~\citep{Anthropic2025Claude}, legal services~\citep{harveyBuildingBusinessCase2022}, financial analysis~\citep{endexAIBuiltExcel2026}, and scientific discovery~\citep{lu_towards_2026}. As these systems proliferate, benchmarks have become the standard tool for comparing performance across vendors and over time. Most benchmarks distill performance into a single headline metric: overall accuracy, defined as the proportion of tasks an agent solves correctly. This metric has shown steady improvement for years, but on many widely used benchmarks, progress has begun to plateau. Top agents now cluster near ceiling-level scores and are often statistically indistinguishable from one another~\citep{akhtarWhenAIBenchmarks2026, jimenez2024swebench, ARCAGI1, chen2021evaluating, hendrycks2021measuringiclr}. Many in the field interpret this as evidence that such benchmarks have lost their discriminative power. The prevailing response has been to retire these benchmarks in favor of more difficult successors; for example, ARC-AGI 1 progressing to ARC-AGI 2 and 3, MMLU to MMLU-Pro, HumanEval to HumanEval+, and SWE-bench to SWE-bench Pro~\citep{CholletARCAGI2NewChallenge2026, foundationARCAGI3NewChallenge2026, wang2024mmlu, liu2023your, dengSWEBenchProCan2025}. We refer to this recurring pattern as \emph{retire-and-replace}.

In this paper, we argue that although this strategy may be useful for model developers who focus on optimizing relative accuracy, it is fundamentally inadequate for helping researchers and downstream developers understand how well an agent solves a real-world task. A central thesis of our work is that \emph{accuracy saturation}, i.e., the state in which top-performing agents achieve statistically indistinguishable accuracies, does not imply that there exist no further insights into performance across all meaningful dimensions of agent behavior. We demonstrate that even when a benchmark’s accuracy metrics have plateaued, we can obtain useful information on agent performance along other critical axes. These include (i) \emph{benchmark validity}, i.e., whether high scores reflect genuine task mastery rather than exploited shortcuts or overfitting; (ii) \emph{evaluation completeness}, capturing reliability, computational efficiency, and the relative performance of the model versus its scaffolding; and (iii) the \emph{practical impact on human workflows}. Although prior work has advocated for evaluating these multifaceted dimensions in principle \citep{Kapoor2025AI, rabanserScienceAIAgent2026, wangPositionHumansAre2026, liang2023holistictmlr, kapoor2026holistic}, the field has largely defaulted to developing increasingly difficult benchmark successors that continue to optimize solely for accuracy. By challenging the \emph{retire-and-replace} paradigm, we argue for extracting the rich signals that persist beyond a benchmark’s accuracy ceiling and emphasize that the limits of accuracy-centric evaluation are present throughout a benchmark’s lifecycle, not just at saturation.

We study this claim through CORE-Bench Hard~\citep{siegel2024corebenchtmlr}, a benchmark for computational reproducibility. CORE-Bench Hard is a useful case study because reproducibility is a high-value, real-world task with a direct human counterpart (enabling a concrete human uplift study), clear out-of-distribution axes (e.g., changing the research fields), and multiple practically relevant performance dimensions (e.g., correctness, cost, latency, and reliability).\footnote{We provide code, data, and logs here: \url{https://github.com/nnadgi01/corebench-analysis}.} Specifically, we make the following three contributions:

\begin{enumerate}[leftmargin=1.5em, topsep=0pt, itemsep=3pt, parsep=0pt]
    \item \textbf{New CORE-Bench variants to improve benchmark validity (\Cref{sec:validity}).} We use log analysis to uncover 15 task-level errors and 20 tasks with exploitable shortcuts in CORE-Bench Hard that would have been difficult to surface before accuracy saturation. We correct these and add ten new tasks to produce CORE-Bench v1.1, a 39-task suite that preserves CORE-Bench Hard's original disciplines, languages, and construction pipeline. We also test whether saturated accuracy transfers under field distribution shift by introducing CORE-Bench OOD, with 19 tasks covering different disciplines from CORE-Bench Hard: physics, engineering, economics, and computer science. We provide a description of each CORE-Bench variant in \Cref{table:core_versions}.
    \item \textbf{Results from multidimensional evaluation (\Cref{sec:multi-metric}).} Even after a benchmark loses discriminative power w.r.t. agent accuracy, it remains useful for differentiating performance across other dimensions. Across 20 agent runs, we show that agents with statistically indistinguishable accuracies differ in efficiency, reliability, and model–scaffold behavior.
    \item \textbf{Observations from measuring the uplift of agent collaborators on human performance (\Cref{sec:uplift}).} While benchmarks are useful proxies for agent capability in task automation, they are insufficient indicators of practical utility for human-agent collaboration. We run a small randomized study on real-world computational reproducibility tasks comprising 20 machine learning and social science papers, and find that agent collaboration more than \textit{halves} completion time. This is likely a conservative estimate, since one-fifth of human-only sessions never completed before reaching the three-hour time limit while all human-agent collaborative sessions did.
\end{enumerate}

\begin{table}[t]
\vspace{-9pt}
\begin{minipage}{\textwidth}
    \centering
    \caption{\textbf{CORE-Bench variants.} CORE-Bench v1.1 corrects threats to construct validity in CORE-Bench Hard. CORE-Bench OOD is an out-of-distribution task suite of CORE-Bench v1.1.}
    \label{table:core_versions}
    \smallskip
    \adjustbox{width=\linewidth}{
    \footnotesize
    \begin{tabular}{@{}p{2.7cm}p{11cm}@{}}
    \toprule
    \textbf{CORE-Bench variant} & \textbf{Description} \\
    \midrule
    CORE-Bench & Original CORE-Bench variant \citep{siegel2024corebenchtmlr} that evaluates agents on computational reproducibility tasks across three fields (computer science, medical science, and the social sciences) and two languages (Python and R). The test set consists of 45 tasks at each of three difficulty levels: Easy, Medium, and Hard. Each task is selected from a capsule on \url{codeocean.com} that contains the codebase of a research paper that is verified to be locally reproducible.\footnote{A capsule is a self-contained, executable research environment that bundles code, data, and software dependencies needed to reproduce a computational experiment.} Each capsule corresponds to a single task, and a task is made up of one or more task questions. While task questions are identical across the three difficulty levels, the agent is provided with less information about solving the task as the difficulty level increases. We refer to \citet{siegel2024corebenchtmlr} for the full capsule-selection criteria.\\ 
    \midrule
    CORE-Bench Hard & Most difficult level of CORE-Bench, where agents must reproduce a paper's code given only the README, the code, and the data (no Dockerfile, runfile, or other instructions). \\
    \midrule
    CORE-Bench v1.1 & Updated version of CORE-Bench Hard. Corrects the 15 task-level errors (spanning incorrect ground truths, malformed task questions, grading errors, and unsolvable tasks) and 20 tasks that allow shortcuts in CORE-Bench Hard. Includes 10 new tasks created using the same construction process and task distribution as CORE-Bench Hard, for 39 total tasks. \\
    \midrule
    CORE-Bench OOD & Suite of 19 tasks designed to evaluate agent performance across a field distribution shift from the other CORE-Bench variants, which consist only of tasks from computer science, medical science, and the social sciences. CORE-Bench OOD evaluates generalizability across fields by covering physics, engineering, economics, and computer science tasks. \\
    \bottomrule
    \end{tabular}
    }
\end{minipage}
\end{table}

\section{Accuracy saturation surfaces threats to benchmark validity}
\label{sec:validity}
Benchmark validity is threatened along two axes that are difficult to anticipate during construction. The first is \emph{task-level threats}, where headline metrics do not faithfully measure the intended capability. Recent work has shown that this affects many widely used benchmarks, including SWE-Bench tasks that are impossible to solve \citep{jimenez2024swebench, ChowdhuryNeil24Introducing}, a $\tau$-Bench Airline scaffold bug \citep{yao2025taubench, kapoor2026holistic}, and WebArena tasks with incorrect grading \citep{zhou2024webarena, Zhu2025Establishing}. These surface once more capable agents are able to exploit alternative solution paths, uncover subtle shortcuts, or succeed end-to-end but are graded incorrectly. \Cref{table:log_analysis} illustrates examples of task-level threats in CORE-Bench Hard that surfaced only once accuracy saturated. Log analysis, the tracking of an agent's inputs, outputs, and environment, has emerged as a key method for identifying them \citep{ukaisiPipelineTranscriptAnalysis2026}, and prior work has used it to uncover benchmark bugs, shortcuts, environmental barriers, and scaffold-level errors \citep{CheatingAIAgent2025, MALTDatasetNatural2025, kapoor2026holistic}. 

The second is \emph{benchmark-specific adaptation}, which arises when benchmarks are used as development targets: as developers iterate on agents against a fixed benchmark, they may adjust prompts, scaffolds, tool-use, dependency handling, timeout settings, or recovery heuristics based on observed failures. These changes improve benchmark performance but can also tailor the agent to the benchmark's idiosyncrasies (e.g., task distributions or output formats). Hence, strong performance may partly reflect adaptation rather than general capability \citep{Kapoor2025AI}.

Accuracy saturation (as defined by \citet{akhtarWhenAIBenchmarks2026}) enables deeper investigation of benchmark validity along both axes. Motivated by this, we introduce two new task suites: CORE-Bench v1.1, which improves construct validity relative to CORE-Bench Hard, and CORE-Bench OOD, which evaluates out-of-distribution generalization.

\subsection{CORE-Bench v1.1: A more robust measure of computational reproducibility}
\label{sec:updated_benchmark}

\begin{wraptable}{r}{0.56\textwidth}
\vspace{-9pt}
\caption{\textbf{CORE-Bench v1.1 accuracies.} Top  agents converge at near-ceiling accuracies. For Claude models, ``thinking'' denotes reasoning (10K token budget for Opus 4.5; "adaptive" has no budget parameter). \texttt{max\_thr} controls the maximum number of concurrent Codex CLI subagents (omitting it disables subagents). Accuracies shown as $\text{value}^{\text{upper}}_{\text{lower}}$ with 95\% Wilson CI bounds.}
\label{table:main_accuracy}
\centering
\scriptsize
\providecommand{\bnd}[1]{\text{\fontsize{4pt}{4.5pt}\selectfont #1}}
\providecommand{\ci}[2]{^{\raisebox{-1pt}{$\bnd{#1}$}}_{\raisebox{1pt}{$\bnd{#2}$}}}
\renewcommand{\arraystretch}{1.1}
\begin{tabular}{llc}
\toprule
\textbf{Scaffold} & \textbf{Model (reasoning effort)} & \textbf{Accuracy}\\
\midrule
\multirow{7}{*}{Codex CLI (default)} & GPT-5 (\texttt{medium}) & $84.6\%\ci{92.8}{70.3}$ \\
                                       & GPT-5.1 (\texttt{medium}) & $87.2\%\ci{94.4}{73.3}$ \\
                                       & GPT-5.2 (\texttt{medium}) & $94.9\%\ci{98.6}{83.1}$ \\
                                       & GPT-5.3-Codex (\texttt{medium}) & $97.4\%\ci{99.5}{86.8}$ \\
                                       & GPT-5.4 (\texttt{low}) & $92.3\%\ci{97.3}{79.7}$ \\
                                       & GPT-5.4 (\texttt{medium}) & $94.9\%\ci{98.6}{83.1}$ \\
                                       & GPT-5.4 (\texttt{high}) & $97.4\%\ci{99.5}{86.8}$ \\
                                       & GPT-5.4 (\texttt{xhigh}) & $97.4\%\ci{99.5}{86.8}$ \\
\midrule
Codex CLI (\texttt{max\_thr=1}) & GPT-5.4 (\texttt{medium}) & $94.9\%\ci{98.6}{83.1}$ \\
Codex CLI (\texttt{max\_thr=3}) & GPT-5.4 (\texttt{medium}) & $97.4\%\ci{99.5}{86.8}$ \\
Codex CLI (\texttt{max\_thr=6}) & GPT-5.4 (\texttt{medium}) & $92.3\%\ci{97.3}{79.7}$ \\
Codex CLI (\texttt{max\_thr=9}) & GPT-5.4 (\texttt{medium}) & $97.4\%\ci{99.5}{86.8}$ \\
\midrule
\multirow{2}{*}{Claude Code} & Opus 4.5 (thinking) & $89.7\%\ci{95.9}{76.4}$ \\
                                       & Opus 4.6 (adaptive) & $92.3\%\ci{97.3}{79.7}$ \\
\midrule
\multirow{3}{*}{OpenCode} & Opus 4.5 (thinking) & $82.1\%\ci{91.0}{67.3}$ \\
                                       & Opus 4.6 (\texttt{none}) & $82.1\%\ci{91.0}{67.3}$ \\
                                       & GPT-5.4 (\texttt{high}) & $84.6\%\ci{92.8}{70.3}$ \\
\midrule
\multirow{3}{*}{CORE-Agent} & Opus 4.5 (\texttt{none}) & $82.1\%\ci{91.0}{67.3}$ \\
                                       & Opus 4.6 (\texttt{none}) & $100\%\ci{100}{91.0}$ \\
                                       & GPT-5.4 (\texttt{medium}) & $51.3\%\ci{66.1}{36.2}$ \\
\bottomrule
\end{tabular}
\vspace{-10pt}
\end{wraptable}

We introduce \emph{CORE-Bench v1.1}, a corrected benchmark developed by identifying task-level threats to construct validity in CORE-Bench Hard via log analysis. 
We construct CORE-Bench v1.1 by applying automated and manual log analysis to the 45 original CORE-Bench Hard tasks and 27 new candidate tasks created for the AgentBeats competition~\citep{berkeley_rdi_agentx_2026}. Rather than serving as a novel, more difficult benchmark, CORE-Bench v1.1 repurposes CORE-Bench Hard. We inspect trajectories using Docent~\citep{meng2025docent} for process correctness, computation correctness, pre-existing artifact contamination, and grading errors using the rubrics in \Cref{table:docent_rubrics}. This process removes or leads to edits for tasks with threats to construct validity; these were difficult to surface prior to accuracy saturation, since less capable agents were not progressing far enough to exploit shortcuts or encounter errors. It yields a final 39-task benchmark: 13 computer science, 10 social science, and 16 medical science tasks. Full construction details are in \Cref{appendix:all_changes}.
We provide a visual overview of the benchmark construction process in \Cref{figure:pipeline}. 

\textbf{Results.} Nicholas Carlini submitted a Claude Code scaffold that obtained a near-ceiling accuracy on CORE-Bench Hard after manually correcting a few grading errors. Despite the construct validity improvements introduced in CORE-Bench v1.1, \emph{accuracy saturation persists}: the top agent obtains an accuracy of 100\% and the next four agents tie at 97.4\% (see \Cref{table:main_accuracy}), so accuracy alone no longer distinguishes leading agents. At the same time, our results also highlight the importance of the scaffold, a finding we discuss in more detail in \Cref{sec:trajectory}. For example, with GPT-5.4 (medium), Codex CLI outperforms the CORE-Agent scaffold by $\approx$ 44 pp.

\begin{table}[t]
\vspace{-9pt}
    \centering
    \caption{We analyzed the logs of our top-performing agents using Docent, an online tool that uses language models to automatically flag an agent's actions from its logs based on a pre-defined rubric \citep{meng2025docent}. Our rubrics were designed to surface threats to construct validity that could either lead to underestimation or overestimation of agent capabilities. We supplemented automated log analysis with manual log inspection of all incorrect tasks and all tasks flagged by the rubric across runs. We conducted log analysis using GPT-5 with medium reasoning and GPT-5.4 with low reasoning.}
    \label{table:docent_rubrics}
    \smallskip
    \adjustbox{width=\linewidth}{
    \footnotesize
    \begin{tabular}{@{}p{3cm}p{10cm}@{}}
    \toprule
    \textbf{For tasks graded as:} & \textbf{We inspect logs to see whether:} \\
    \midrule
    Incorrect & The agent solves the intended task end-to-end and gives a logically or procedurally correct answer based on the environment or reasoning in the transcript.  \\
    \midrule
    Correct & The agent either doesn't reproduce the paper correctly (process incorrectness) or doesn't perform the correct final computation (computation incorrectness). \\
    \midrule
    All tasks & The agent is able to obtain the correct answer to a task by directly reading a value that already exists (pre-run) inside static artifacts or rendered documents, or applying only extremely trivial operations over values in the pre-existing artifacts (for example, very simple filtering-plus-counting or literal pattern-counting in text). \\
    \midrule
    \end{tabular}
    }
\end{table}

\subsection{CORE-Bench OOD: An out-of-distribution task suite of CORE-Bench v1.1}
\label{sec:ood}

\emph{CORE-Bench OOD} tests whether performance on CORE-Bench v1.1 transfers under a field distribution shift. This shift is critical, as disciplines vary significantly in repository organization, software ecosystems, manuscript conventions, and computational workflows. While preserving the underlying task structure of v1.1, CORE-Bench OOD changes the disciplinary composition as follows: two economics, ten engineering, five physics, and two computer science tasks (one of which has a runtime of around 50 minutes). Following the same log analysis procedures used for v1.1 (see \Cref{sec:updated_benchmark}), we evaluated an initial pool of 30 OOD tasks written at the same time as CORE-Bench Hard using CORE-Agent (Opus 4.5 and 4.6) and OpenCode (GPT-5.2). This initial round of removing 12 tasks, editing 8, and adding 6 yielded a 24-task subset. Subsequent log analysis of incorrect tasks across 12 Codex CLI runs identified further errors, prompting the removal of 5 additional tasks and the regrading of one to establish the final 19-task benchmark (see \Cref{fig:pipeline_ood} and \Cref{appendix:all_changes} for details). 

\begin{wraptable}{r}{0.56\textwidth}
\vspace{-9pt}
\caption{\textbf{CORE-Bench OOD accuracies.} The top five of 12 Codex CLI agents (varying model, reasoning effort, and \texttt{max\_thr}) cluster at near-ceiling, statistically indistinguishable accuracies (\Cref{appendix:saturation_section}). Accuracies shown as $\text{value}^{\text{upper}}_{\text{lower}}$ with 95\% Wilson CI bounds.}
\label{table:ood_accuracy}
\centering
\scriptsize
\providecommand{\bnd}[1]{\text{\fontsize{4pt}{4.5pt}\selectfont #1}}
\providecommand{\ci}[2]{^{\raisebox{-1pt}{$\bnd{#1}$}}_{\raisebox{1pt}{$\bnd{#2}$}}}
\renewcommand{\arraystretch}{1.13}
\begin{tabular}{llc}
\toprule
\textbf{Scaffold} & \textbf{Model (reasoning effort)} & \textbf{Accuracy}\\
\midrule
\multirow{7}{*}{Codex CLI (default)} & GPT-5 (\texttt{medium}) & $89.5\%\ci{97.1}{68.6}$ \\
                                       & GPT-5.1 (\texttt{medium}) & $94.7\%\ci{99.1}{75.4}$ \\
                                       & GPT-5.2 (\texttt{medium}) & $100.0\%\ci{100.0}{83.2}$ \\
                                       & GPT-5.3-Codex (\texttt{medium}) & $89.5\%\ci{97.1}{68.6}$ \\
                                       & GPT-5.4 (\texttt{low}) & $84.2\%\ci{94.5}{62.4}$ \\
                                       & GPT-5.4 (\texttt{medium}) & $89.5\%\ci{97.1}{68.6}$ \\
                                       & GPT-5.4 (\texttt{high}) & $89.5\%\ci{97.1}{68.6}$ \\
                                       & GPT-5.4 (\texttt{xhigh}) & $100.0\%\ci{100.0}{83.2}$ \\
\midrule
Codex CLI (\texttt{max\_thr=1}) & GPT-5.4 (\texttt{medium}) & $94.7\%\ci{99.1}{75.4}$ \\
Codex CLI (\texttt{max\_thr=3}) & GPT-5.4 (\texttt{medium}) & $89.5\%\ci{97.1}{68.6}$ \\
Codex CLI (\texttt{max\_thr=6}) & GPT-5.4 (\texttt{medium}) & $84.2\%\ci{94.5}{62.4}$ \\
Codex CLI (\texttt{max\_thr=9}) & GPT-5.4 (\texttt{medium}) & $84.2\%\ci{94.5}{62.4}$ \\
\bottomrule
\end{tabular}
\vspace{-10pt}
\end{wraptable}

\textbf{Results.} We evaluate 12 Codex CLI agents on CORE-Bench OOD, varying the model, reasoning effort, and number of subagents invoked. We present results in \Cref{table:ood_accuracy}, and we find that the top five agents obtain statistically indistinguishable accuracies on CORE-Bench OOD, indicating that \emph{accuracy saturation on CORE-Bench v1.1 translates across a discipline distribution shift}.

We further note that log analysis is not exhaustive: it requires specifying target behaviors, some threats to construct validity surface only in particular agent runs, and LLM-based classifiers require manual validation. We therefore treat CORE-Bench v1.1 and CORE-Bench OOD as active benchmarks that we plan to update as new validity threats are found.

\newpage
\section{Multidimensional evaluation of agent performance}
\label{sec:multi-metric}

\begin{wraptable}[21]{r}{0.52\columnwidth}
\vspace{-1.5pt}
\centering
\caption{\textbf{Root-cause taxonomy of 56 accuracy failures.} Failure
modes are unevenly distributed across scaffolds: wrong-metric errors
concentrate in CORE-Agent, while timeouts and dependency failures
concentrate in OpenCode. ``Spiraling'' timeouts reflect repeated failed
fix attempts; ``environment'' timeouts reflect slow-running processes.
CC: Claude Code; Cx: Codex CLI; OC: OpenCode; CA: CORE-Agent.}
\label{table:root_cause}
\vspace{4pt}
\footnotesize
\setlength{\tabcolsep}{4pt}
\renewcommand{\arraystretch}{1.1}
\begin{tabular}{@{}lcccc|c@{}}
\toprule
\textbf{Failure Root cause} & \textbf{CC} & \textbf{Cx} & \textbf{OC} & \textbf{CA} & \textbf{Total} \\
\midrule
    Wrong metric / computation   & 2 & 2 & 0  & 14 & 18 \\
    Timeout (spiraling on fixes) & 3 & 0 & 8  & 3  & 14 \\
    Gave up (no answer)          & 0 & 0 & 5  & 2  & 7  \\
    Dependency failure           & 0 & 0 & 6  & 0  & 6  \\
    Vision / web fallback        & 0 & 0 & 0  & 5  & 5  \\
    Precision / rounding         & 0 & 0 & 0  & 2  & 2  \\
    Timeout (environment)        & 2 & 0 & 1  & 0  & 3  \\
    Format mismatch              & 0 & 1 & 0  & 0  & 1  \\
\midrule
\textbf{Total failures}          & 7 & 3 & 20 & 26 & 56 \\
\bottomrule
\end{tabular}
\label{tab:root_cause}
\end{wraptable}

Our accuracy saturation results on CORE-Bench v1.1 and CORE-Bench OOD limit the usefulness of both benchmarks for distinguishing between agents by accuracy. Consequently, we propose decoupling \textit{accuracy saturation} from \textit{benchmark saturation}: we show that extending a benchmark's lifecycle beyond accuracy to measuring additional dimensions of agent performance (reliability, efficiency, and the relative importance of the model versus the scaffold) retains their utility as a proxy for agent performance even after accuracy saturates. While accuracy-centric evaluations are insufficient measurement tools even before accuracy saturates, saturation highlights the immediate necessity of moving beyond them.

\subsection{Reliability}
\label{sec:reliability}

\begin{figure*}[t]
  \centering
  \begin{minipage}[c]{0.65\textwidth}
    \centering
    \begin{subfigure}[t]{0.485\linewidth}
      \centering
      \includegraphics[width=\linewidth, height=10.35em]{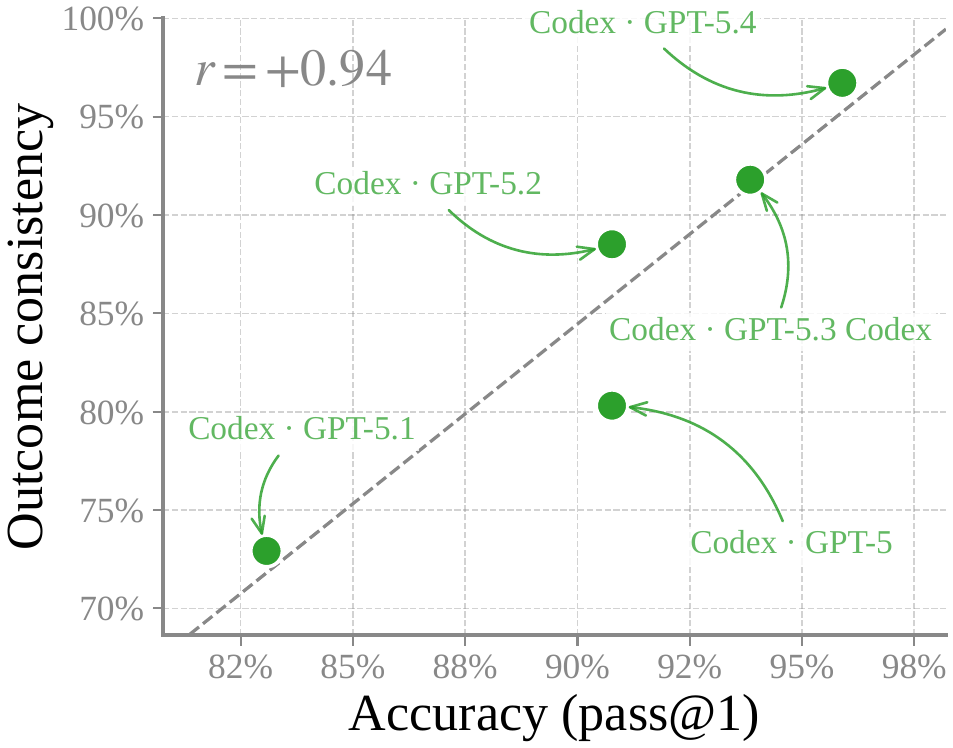}
      \caption{More accurate agents have more consistent outputs across runs.}
      \label{fig:reliability-outcome}
    \end{subfigure}
    \hfill
    \begin{subfigure}[t]{0.485\linewidth}
      \centering
      \includegraphics[width=\linewidth, height=10.35em]{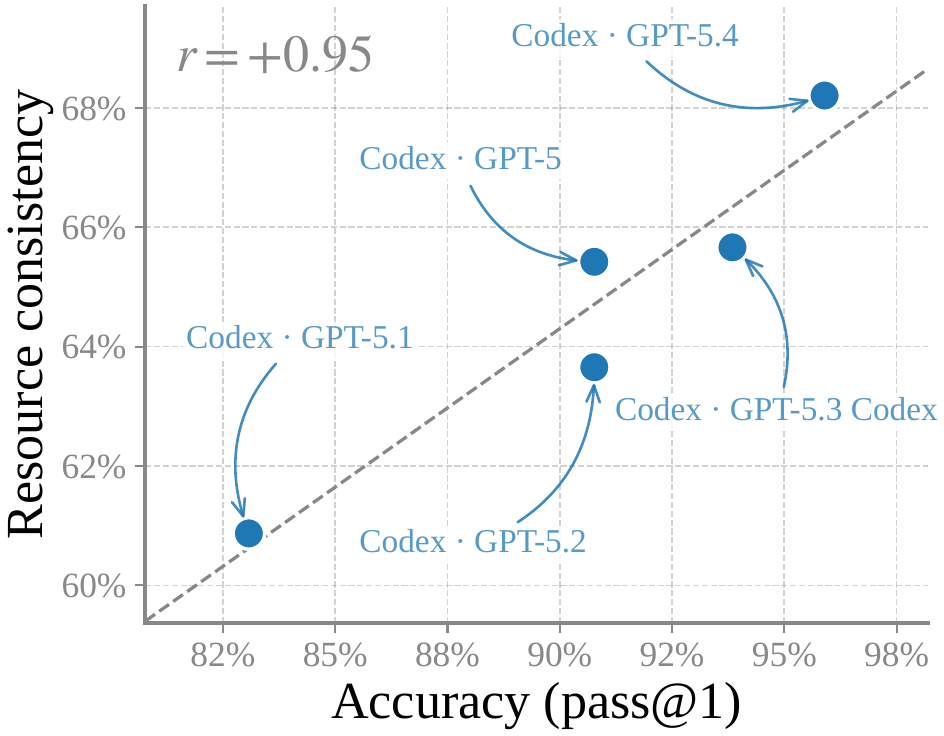}
      \caption{More accurate agents use a more consistent \# of tokens per run.}
      \label{fig:reliability-resource}
    \end{subfigure}

    \vspace{1.24em}

    \begin{subfigure}[t]{0.485\linewidth}
      \centering
      \includegraphics[width=\linewidth, height=10.35em]{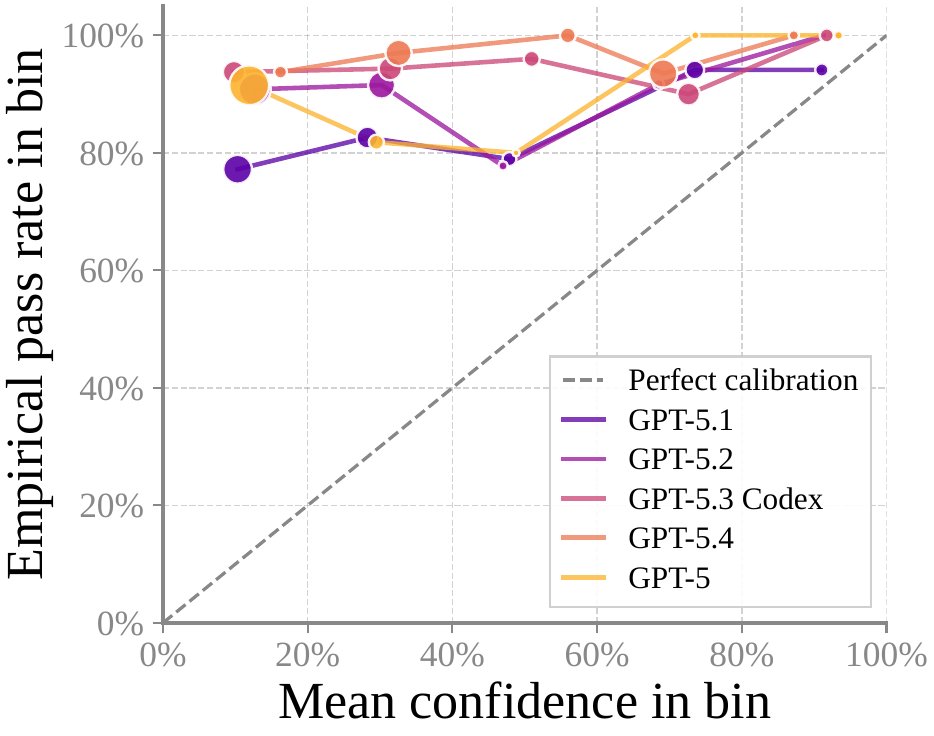}
      \caption{Agents are poorly calibrated, generally being under-confident.}
      \label{fig:tokens-vs-accuracy}
    \end{subfigure}
    \hfill
    \begin{subfigure}[t]{0.485\linewidth}
      \centering
      \includegraphics[width=\linewidth, height=10.35em]{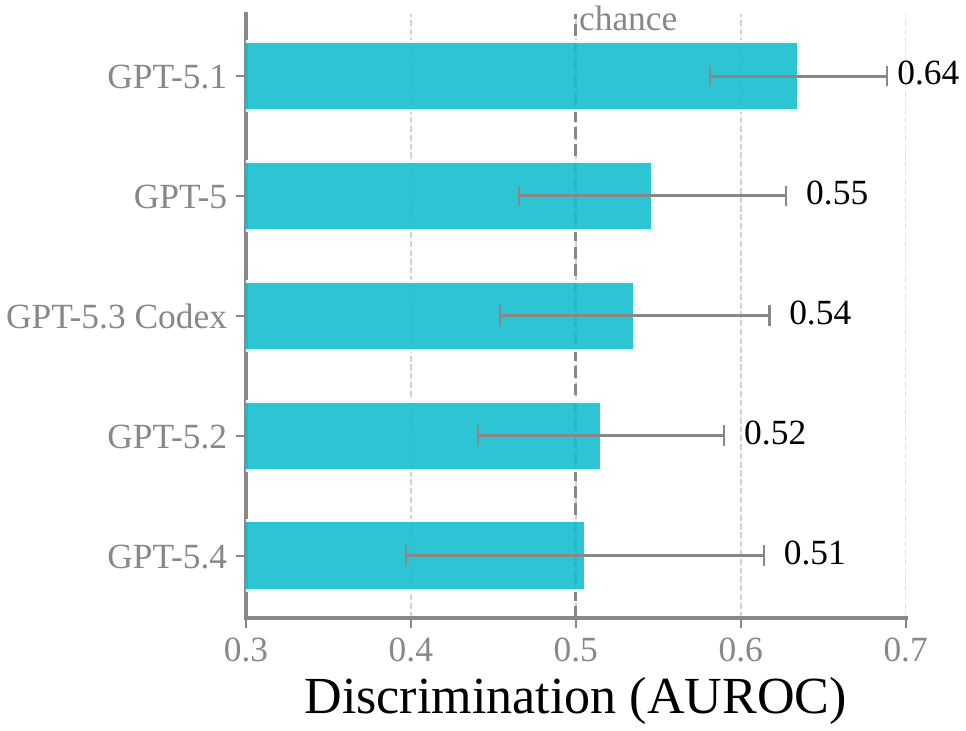}
      \caption{Agents are not able to distinguish successes from failures.}
      \label{fig:cost-vs-accuracy}
    \end{subfigure}
  \end{minipage}%
  \hfill
  \begin{minipage}[c]{0.32\textwidth}
    \centering
    \begin{subfigure}[t]{\linewidth}
      \centering
      \includegraphics[
        width=\linewidth,
        height=8cm,
        trim={0 0 0 0},
        clip
      ]{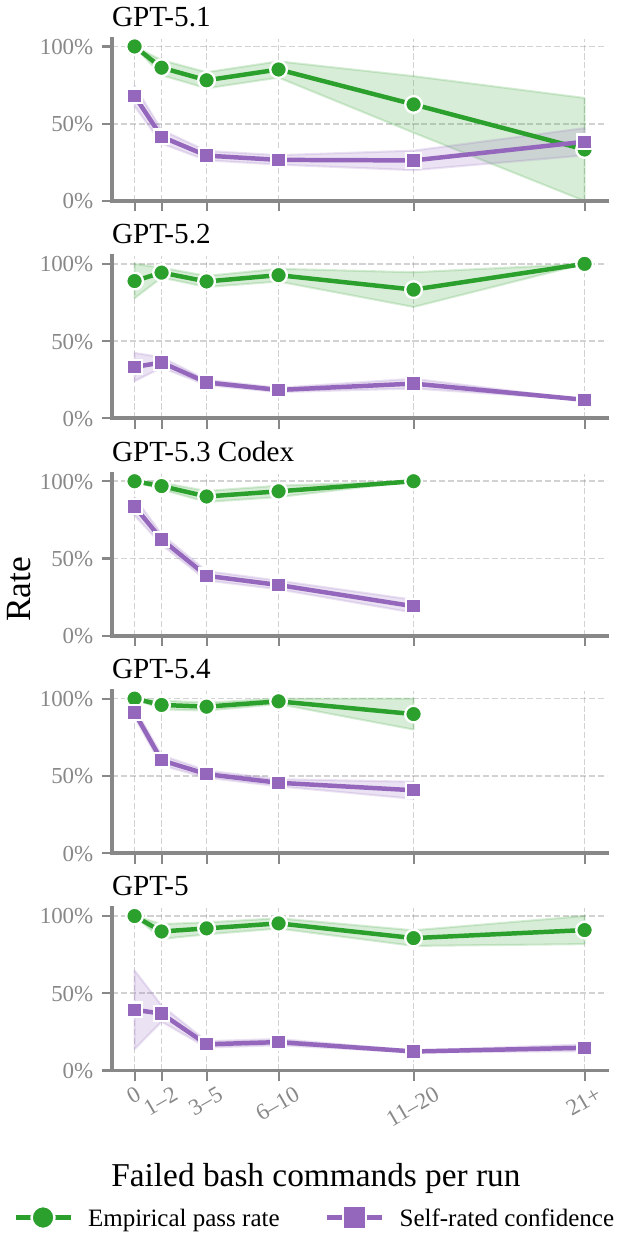}
      \caption{Agents are broadly underconfident; their confidence tracks failed bash commands per task, a metric uncorrelated with task success.}
      \label{fig:reliability-predictability}
    \end{subfigure}
  \end{minipage}
  \caption{
  Reliability analyses.
  \textbf{(a)} Outcome consistency and \textbf{(b)} resource consistency
  both increase with reliability-sample accuracy, indicating that more
  accurate agents are also more repeatable across runs.
  \textbf{(c)} Agents are systematically underconfident and \textbf{(d)} frequently do not exhibit discrimination better than random chance. 
  \textbf{(e)} Per-agent predictability curves: empirical pass rates
  remain high across tool-error bins, while self-rated confidence declines with failed bash commands.
  }
  \label{fig:reliability-results}
\end{figure*}

Two agents with identical mean accuracy can differ substantially in how consistent their outputs are across repeated runs and in how well their stated confidence anticipates success. We adopt the reliability framework of \citet{rabanserScienceAIAgent2026} and measure four metrics: \emph{outcome consistency} (rate at which repeated runs of a task yield the same verdict), \emph{resource consistency} (variability in tokens), \emph{calibration} (gap between stated confidence and empirical success), and \emph{discrimination} (whether confidence rankings separate successes from failures).\footnote{The framework also covers robustness and safety, which we address via the OOD analysis (\Cref{sec:ood}) and benchmark validity analysis (\Cref{sec:validity}).}
We run five additional trials on each of five Codex CLI agents (GPT-5, GPT-5.1, GPT-5.2, GPT-5.3-Codex, and GPT-5.4, all at medium reasoning), eliciting post-hoc confidence via an additional prompt at the end of the agent interaction.\footnote{We use Codex CLI v0.122 for GPT-5.1 and Codex CLI v0.130.0 for all other models. See \Cref{sec:codex_version} for details.} Based on our results from \Cref{fig:reliability-results}, we draw the following key conclusions: 
\begin{enumerate}[leftmargin=1.5em]
    \item \textbf{In a small sample, agents that are more accurate on average are also more consistent.} The most accurate agent based on the average score across five runs also has the most consistent outputs (dependably correct or incorrect) and the most consistent token usage.
    \item \textbf{Agents are massively under-confident and struggle to separate correct from incorrect responses.} While the mean empirical pass rate across all runs is 93\%, the mean reported confidence is only 32.1\%. Reported confidence tracks the number of bash tool errors, but this metric is uncorrelated with task success. In fact, no agent appears to be outperforming a simple random guessing baseline telling correct and incorrect tasks apart based on confidence.
\end{enumerate}

\subsection{Efficiency}
\label{sec:efficiency}

\begin{figure*}[t]
  \centering
  \includegraphics[width=\linewidth]{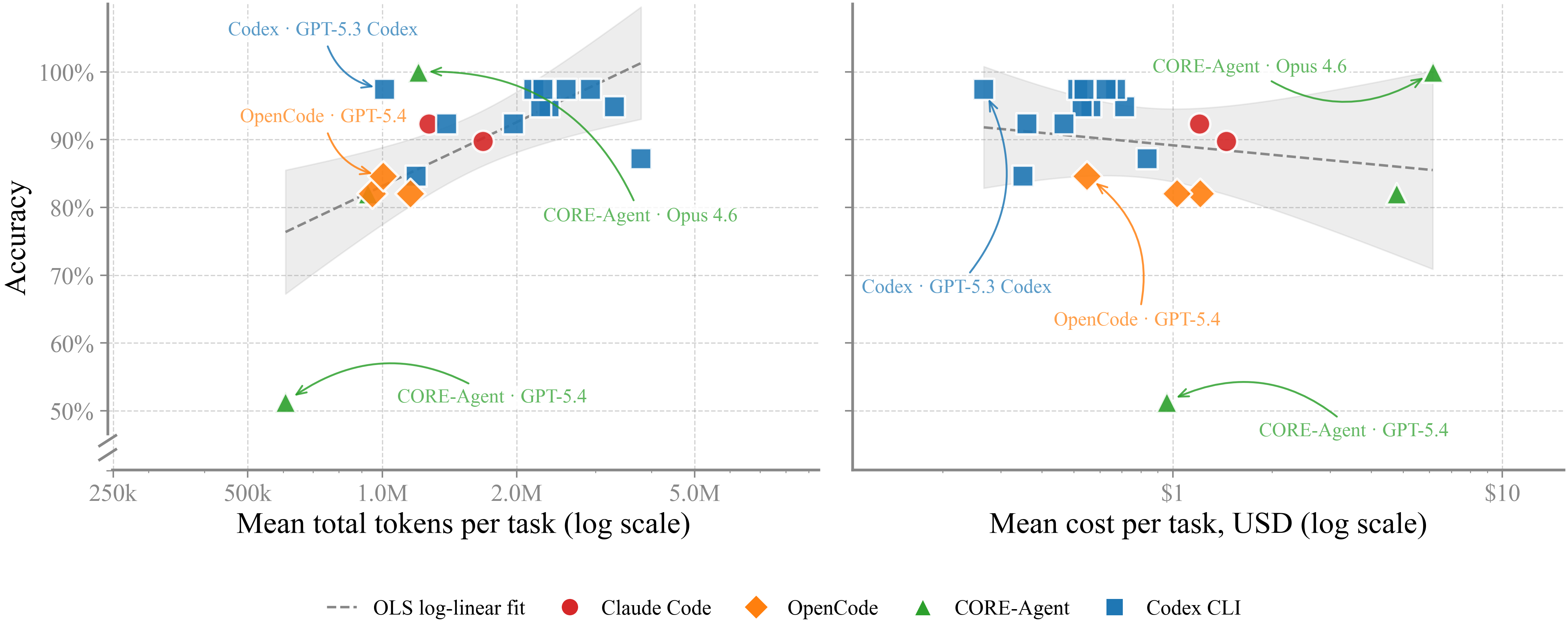}
  \caption{
    \textbf{Efficiency measured by accuracy vs.\ total token usage and estimated cost.} GPT-5.3-Codex is the most efficient high-accuracy agent by both token usage and cost. The relationship between token usage and accuracy is not reflected between cost and accuracy.}
  \label{fig:efficiency-2d}
\end{figure*}

The well-documented returns from inference scaling \citep{brown2024large, guo2025deepseekr1} show that agents can often achieve high accuracy simply by using more compute. For researchers, this "brute force" capability is useful for identifying the upper bounds of a model's potential. However, for most practitioners, the cost of reaching an answer is just as important as the answer itself. To address this, we analyze efficiency by measuring both token usage and total dollar cost.\footnote{For CORE-Agent with Opus 4.5, we drop two tasks from the mean resource usage calculation. These tasks timed out, so resource usage was not logged.} Token usage includes the sum of all input, cached, and output tokens, while the dollar cost is calculated based on prices at the time of the run. In Figure \ref{fig:efficiency-2d}, we plot accuracy against these two metrics. From this data, we highlight two main findings:
\newpage

\begin{enumerate}[leftmargin=1.5em]
    \item \textbf{Some high-scoring agents are much more efficient than others.} Cost-aware analysis allows us to differentiate between our top scoring agents. GPT-5.3-Codex (medium) is most efficient by both token usage and cost. Compared to GPT-5.4 (high), which achieved equal accuracy (97.4\%), GPT-5.3-Codex (medium) costs roughly 60\% less. 
    \item \textbf{Token usage and cost tell different stories of efficiency.} Token usage and cost have different relationships with accuracy. This is principally driven by model provider pricing, some Codex CLI model-scaffold pairs caching most aggressively, and CORE-Agent not caching at all.
\end{enumerate} 

\subsection{Decoupling model and scaffold}
\label{sec:trajectory}

Agent benchmark leaderboards typically report a single accuracy per agent, collapsing the contributions of the underlying model and the scaffold that orchestrates it. When accuracy improves from one leaderboard entry to the next, it is therefore unclear whether the gain is attributable to a more capable model, a better-engineered scaffold, or a better match between the two. Accuracy saturation makes this question more important: once several agents reach statistically similar top-line accuracy, the leaderboard no longer reveals which part of the agent stack is responsible for success. 

Our evaluation design provides model-scaffold comparisons that allow us to probe these effects. We evaluate Opus 4.5, Opus 4.6, and GPT-5.4, on three of four scaffolds each. Claude Code is a proprietary, vendor-developed scaffold. CORE-Agent (built on HuggingFace smolagents), OpenCode, and Codex CLI are open-source scaffolds. We provide scaffold configurations in \Cref{appendix:core_implementation} and reasoning configurations vary as per \Cref{table:main_accuracy}. We inspected the trajectories of tasks where outcomes varied across models and scaffolds, classified all 56 failures by root cause using Docent (GPT-5.5 and high reasoning), and applied a Docent rubric to all 390 logs to surface trajectory differences. 

\textbf{Results.} Our analysis reveals three findings:

\begin{enumerate}[leftmargin=*,itemsep=0.4em]

\item \textbf{Similar accuracies can mask fundamentally different failures.} We provide representative examples of these disagreements in \Cref{table:root_cause} and \Cref{table:rep_trajectory}. This effect persists even when comparing different scaffolds paired with the same model. For example, Opus 4.5 achieves 82.1\% accuracy on both CORE-Agent and OpenCode, yet the two scaffolds' outcomes disagree on 12 of 39, or 31\% of capsules (see \Cref{fig:passfail_scaffold}). An oracle router that selects the best scaffold per task achieves 100\% accuracy for both Opus 4.5 and GPT-5.4, implying that every task in CORE-Bench v1.1 is solvable by at least one scaffold for these models. This complementarity suggests that scaffolds are altering which tasks models can solve and how they solve them.

\item \textbf{Scaffolds induce distinct solution strategies.} Holding the model constant and swapping only the scaffold makes the scaffold-induced differences visible. Some vision tasks can be solved correctly using code output and without rendering the figure. With Opus 4.6, Claude Code derives 41\% of answers from the text output of unmodified code (no vision-read) and only 3\% from a vision-reading of a rendered figure; the same model with CORE-Agent derives the answer from the text output of unmodified code in just 21\% of runs and reaches for a vision-read 31\% of the time. The pattern sharpens on the other two models: vision-read rates jump from 3\% (Claude Code) to 62\% (CORE-Agent) on Opus 4.5 and from 1\% (Codex CLI) to 56\% (CORE-Agent) on GPT-5.4. Vision-reads following a clean run pass 93\% of the time, but those used as a fallback after the agent abandons the original code (47\%) or gives up entirely (60\%) roughly pass only 50\% of the time. CORE-Agent's accuracy gap is largely the accumulation of these fallback failures.

\item \textbf{Direct fixes strongly outperform rewrites.} Scaffolds that diagnose a root cause and apply a targeted fix succeed 95.2\% of the time ($n=269$), whereas scaffolds that abandon the original implementation and rewrite from scratch succeed only 67.8\% of the time ($n=59$). Restricting the analysis to the 26 capsules where both strategies were attempted shows a similar pattern: 96\% success for direct fixes versus 68\% for rewrites, despite small per-capsule sample sizes. Notably, a scaffold's tendency toward direct fixes closely tracks its overall accuracy: Codex CLI uses direct fixes 82\% of the time, while CORE-Agent does so only 49\% of the time.
\end{enumerate}

Together, these findings suggest that model and scaffold effects are not cleanly separable: scaffolds constrain available solution paths, while models determine how effectively they are used.

\begin{table}[t]
\caption{\textbf{Representative trajectory-level disagreements across scaffolds.}
Each cell summarizes the decisive moment in a model-scaffold-capsule run. The Model provider scaffold column reports Codex CLI for GPT-5.4 runs and Claude Code for Opus runs. We provide specific details on how failure modes differed by task in \Cref{appendix:by_capsule}.}
\label{table:rep_trajectory}
\vspace{10pt}
\centering
\footnotesize
\resizebox{\linewidth}{!}{
\begin{tabular}{@{}p{2.7cm}p{1.4cm}p{3.9cm}p{3.9cm}p{3.9cm}@{}}
\toprule
\textbf{Reproduction target} & \textbf{Model} & \textbf{CORE-Agent} & \textbf{Model provider scaffold} & \textbf{OpenCode} \\
\midrule
{\textbf{capsule-1175539.} \newline
\scriptsize Report the study group with the highest median cardiac concentricity.}
& GPT-5.4
& \textbf{Fail} (8 msgs). Stale environment symlinks and a shallow filesystem search misses the script one directory deeper. Falls back to an unrelated notebook render using a different dataset and extracts the wrong group name from its prose.
& \textbf{Pass} (56 msgs). Permission restrictions block the script's hard-coded absolute paths; after exhausting filesystem workarounds, redirects both input and output paths to the working directory, then runs the original script.
& \textbf{Pass} (35 msgs). Installs dependencies with sudo, runs the script, and computes group medians with a targeted R command to extract the answer. \\
\midrule
\multirow{3}{2.7cm}{\textbf{capsule-4252248.} \newline
\scriptsize Report the PR-curve AUC for the ATC/CHEMBL drug-sensitivity integration benchmark.}
& GPT-5.4
& \textbf{Fail.} (84 msgs). Computes all four benchmark AUCs but selects the sensitivity-layer value instead of the integration-layer value.
& \textbf{Fail.} (110; 138 msgs). High reasoning selects the sensitivity-layer value; medium reasoning skips preprocessing and reports the wrong value.
& \textbf{Pass.} (117 msgs). Bioconductor version conflicts block a required package; rather than resolving the full dependency chain, creates a slim local stand-in that defines only the single class needed to load the data, patches R~4.x bugs, and runs the full pipeline. \\
& Opus 4.5
& \textbf{Fail.} (128 msgs). rJava fails to load despite installing the JDK and reconfiguring Java; falls back to a simplified computation that skips the preprocessing, producing an incorrect value.
& \textbf{Fail.} (180 msgs). Cannot compile R's curl package (missing dev headers for installed libcurl4t64); reimplements the benchmarking pipeline standalone with different preprocessing.
& \textbf{Pass.} (195 msgs). Iteratively resolves Bioconductor version conflicts including a BH downgrade, patches R 4.x compatibility bugs in the paper's code, and runs the full pipeline. \\
\midrule
\multirow{2}{2.7cm}{\textbf{capsule-5136217.} \newline
\scriptsize Recover the Figure 3 political-news sharing result.}
& Opus 4.6
& \textbf{Pass.} (114 msgs). Discovers \texttt{bsts} is unavailable; installs R from scratch and runs the scripts needed for the figure. Reads the answer from the generated plot via a vision model, which returns the wrong group; catches the error by cross-checking against a self-authored Python replication, then verifies via direct R computation.
& \textbf{Pass} (63 msgs). Traces the figure target to two upstream scripts and runs them, and extracts the answer by computing group means directly in R, never rendering or reading the generated plot.
& \textbf{Fail} (32 msgs). Attempts to compile \texttt{Boom} the \texttt{bsts} dependency that isn't needed from source, hitting two bash timeouts, and then times out. \\
& Opus 4.5
& \textbf{Pass} (76 msgs). Creates a modified copy of the relevant script with unavailable packages commented out and runs only what is needed to generate the figure.
& \textbf{Fail} (262 msgs). Largely consumed by dependency installation failures and package workarounds. Derives the correct answer from intermediate data but does not finish before the task timeout elapses.
& \textbf{Fail.} (46 msgs). Compiles \texttt{bsts} successfully, but times out during data processing on the large dataset. \\
\midrule
\multirow{2}{2.7cm}{\textbf{capsule-0851068.} \newline
\scriptsize Reproduce the reported AUC from a PyTorch classification pipeline.}
& GPT-5.4
& \textbf{Fail.} (38 msgs). Data symlinks point to a nonexistent agent run directory, leaving the input folder empty. Rather than repairing the symlinks, searches the web for the paper's reported results and submits an AUC from a different experimental condition.
& \textbf{Pass.} (69 msgs). Runs the demo script; PyTorch's \texttt{DataLoader} crashes because the deep workspace path exceeds the 108-byte \texttt{AF\_UNIX} socket limit at \texttt{num\_workers=16}. Diagnoses the socket-path constraint, patches \texttt{num\_workers=0}, and reruns to completion.
& \textbf{Pass.} (47 msgs).  Hits the same \texttt{AF\_UNIX} socket-path crash, reaches the same diagnosis independently, and applies the same \texttt{num\_workers=0} patch to complete. \\
& Opus 4.6
& \textbf{Pass.} (66 msgs). Discovers that data symlinks point to a different agent run's directory, deletes them, and recreates them against the correct path. After installing PyTorch, proactively reduces \texttt{num\_workers} to 0 before encountering the socket-path error, avoiding the crash entirely.
& \textbf{Fail} (72 msgs). Diagnoses the \texttt{AF\_UNIX} socket-path bug, patches \texttt{num\_workers=0}, and computes the correct AUC, but the 2,700\,s timeout elapses before answer collection.
& \textbf{Pass.} (35 msgs). The most efficient run across both models. Hits the \texttt{AF\_UNIX} error, patches \texttt{num\_workers=0}, and completes in 35 messages. \\
\bottomrule
\end{tabular}
}
\end{table}

\section{Measuring uplift from human-agent collaboration}
\label{sec:uplift}

Real-world computational reproducibility is grounded in scientific workflows where humans interpret, validate, and build on results. Once top-performing agents converge at near-ceiling accuracy, the question shifts from whether agents can complete a task to whether they provide value when deployed alongside humans. High benchmark accuracy may not cleanly translate to uplift: benchmark task distributions might be more limited in scope than real-world tasks, agent failures may be more time-consuming for a human (or the agent itself) to resolve than human failures, or agents may take more time to effectively respond to human redirection. Prior work shows productivity gains from coding agents are highly context-dependent, often emerging only in human-in-the-loop settings \citep{wangPositionHumansAre2026, beckerMeasuringImpactEarly20252025}. To measure this directly, we ran a randomized study in which five evaluators reproduced results from 20 machine learning and social science papers, with and without agent collaboration, to estimate process-level uplift.
\subsection{Methodology}
\label{section: methodology}

\textbf{Paper selection.} We selected 20 papers across machine learning and the social sciences. The machine learning papers were drawn from a list of award-winning papers at major machine learning conferences since 2011 \citep{aibestpapers}. The social science papers were drawn from a dataset published by the Institute for Replication (I4R) \citep{i4rmetadatabase2024}. To enhance representativeness, each selector was given a random subset of papers from each dataset in randomized order and evaluated them sequentially for inclusion according to our paper selection criteria (see Appendix~\ref{appendix:paper_selection} for details) until reaching the required number of selections. Unlike CORE-Bench, the purpose of this study was to gain process-level insights into uplift from human-agent collaboration, rather than validate final answer correctness. Accordingly, the papers were not limited to those that are confirmed to be computationally reproducible. We deliberately included two social science papers that I4R had assessed as not achieving a ``perfect reproduction'' to better reflect real-world computational reproducibility work. A single result from each selected paper was specified as the replication target (see \ref{app:PaperTargets} for a full list).

\textbf{Participants.} Five of the authors joined the experiment as evaluators, all of whom have a master's degree in data science and experience with computational reproducibility tasks. Each of the papers (i.e. reproduction tasks) was independently attempted by two or three of the five evaluators. The same five authors who conducted the replication attempts also carried out paper and reproduction target selection. To ensure blinding, no author was assigned as an evaluator for a paper they had encountered during the selection process. For the social science papers, the selector was aware of whether the paper had been assessed as "perfectly reproducible" by I4R replicators, but the evaluator was not. For all the machine learning papers, the reproducibility status was not known beforehand.

\textbf{Agent configuration.} The human-agent collaboration condition used Codex CLI running GPT-5.4 at the extra-high thinking setting. Participants used a standardized interface but were otherwise free to interact with the agent. We constructed two Docker-based evaluation environments, one for machine learning papers and one for social science papers. Each had tailored Python and R support for replication. We applied a standardized prompt (see \Cref{appendix:default_prompt}) using a fully autonomous execution setting, in which the agent iteratively generates and runs code without intermediate human approval. However, the prompt instructed the agent to stop and escalate to the human when encountering blockers it could not resolve after 2-3 attempts.

\textbf{Paper replication.} We randomly assigned these 20 papers across the five evaluators (see \Cref{appendix:RCTdesign} for details about the randomization design). Each participant attempted 10 papers total, 5 with agent assistance and 5 without, and 5 from each source dataset. Each paper was attempted by two or three participants, with at least one manual and one human-agent collaboration attempt, yielding 50 replication experiments across the 20 papers. To mitigate learning effects, participants were asked to complete tasks in a pre-specified randomized order. In the manual condition (see \Cref{appendix:instructions}), participants were allowed to use traditional web search tools (e.g., Google, StackOverflow) but were prohibited from using generative AI systems or AI search summaries (e.g., ChatGPT, Copilot, or AI overviews in search engines), consistent with prior experimental protocols \citep{beckerMeasuringImpactEarly20252025, hong2026measuringmid2025llmassistancenovice}. In the human-agent collaboration condition, participants applied a shared prompt template (see \Cref{appendix:default_prompt}) uniformly across tasks. We set a maximum time limit of 3 hours per run for both conditions.

\textbf{Task questionnaire.} We adopted design approaches used in prior work to design a questionnaire for documenting agent failure modes and instances of human intervention as structured feedback \citep{paradis2025doesaiimpactdevelopment}. Our questionnaire is provided in \Cref{appendix: Questionnaire}.

\subsection{Results}
\begin{figure}[t]  %
\centering
\begin{minipage}[t]{0.5\linewidth}
    \centering
    \vspace{0pt}  %
    \includegraphics[width=\linewidth]{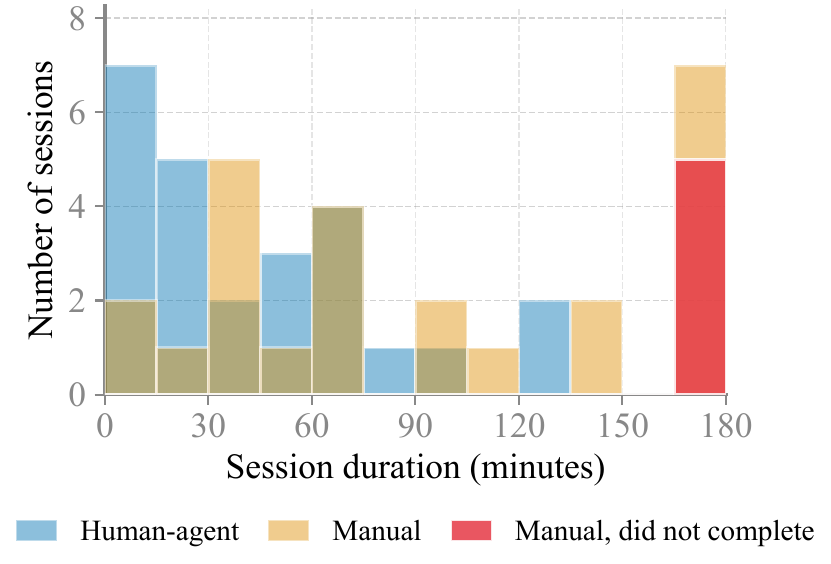}
    \captionof{figure}{\textbf{Distribution of durations of reproduction sessions} in the randomized study for manual vs.\ human-agent collaborative sessions. Evaluators were instructed to abandon runs if no result had been produced yet after three hours, a limit that was only reached during manual sessions.}
    \label{fig:duration_sessions}
\end{minipage}%
\hfill
\begin{minipage}[t]{0.45\linewidth}
    \centering
    \vspace{0pt}
    \label{tab:collaboration_patterns_simple}
    \footnotesize
    \renewcommand{\arraystretch}{1.1}
    \raggedright
    \captionof{table}{\textbf{Observed collaboration patterns across 25 human-agent collaborative reproduction runs.}}
    \vspace{-4pt}
    {\footnotesize\raggedright $^{*}$These observations originated from the same run, and multiple collaboration patterns could be assigned to a single run.\par}
    \begin{tabular}{@{}p{0.82\linewidth}c@{}}
    \toprule
    \textbf{Collaboration pattern observed} & \textbf{Runs} \\
    \midrule
    Agent did all the work on its own & 19 \\
    Minor human suggestions or redirection$^{*}$ & 3 \\
    Agent asked for human input less than 5 times$^{*}$ & 1 \\
    Agent made major error(s), requiring human redirection & 1 \\
    Agent completed task but required significant scope clarification upfront & 1 \\
    Agent wasted a lot of time going down the wrong path, but eventually stopped to check in with the human (as requested in the prompt), to suggest an alternative approach, which worked after human approval & 1 \\
    \bottomrule
    \end{tabular}
\end{minipage}
\end{figure}

Our results show that human-agent collaboration provides substantial uplift on computational reproducibility tasks compared to humans alone. Specifically, we find: 
\begin{enumerate}[leftmargin=*]
    \item \textbf{Human-agent collaboration provides uplift in reproduction time.} Our fixed effects model (see \Cref{appendix:fixed_effects}) estimates that manual reproduction sessions lasted 2.11 times as long as human-agent collaborative sessions. The coefficient estimate's CR2 standard error, clustered by researcher, is 0.09 with a (two-sided) p-value of 0.00176, indicating a statistically significant positive result. The three-hour time limit was reached for five out of 25 manual runs and none of the human-agent runs, suggesting that without this constraint, the estimated uplift would likely be larger (see \Cref{fig:duration_sessions}).
    \item \textbf{Most human-agent collaborative runs required only minimal or no human assistance.} Across 25 human-agent collaborative runs, evaluators reported that the agent was able to complete 19 fully autonomously (aside from two setup steps explicitly assigned to humans: starting the instance and Docker image, and starting the agent). In the remaining six runs, humans intervened mainly during setup, code execution, result comparison, and discrepancy investigation. These interventions ranged from minimal human input to complete redirection (see \Cref{tab:collaboration_patterns_simple} for a full list).
    \item \textbf{Agents were perceived to be the most useful in environment setup and running code.} After each human-agent collaborative reproduction session, the human evaluators were asked to assess "Where Agent [had] added value". The most frequent responses (see \Cref{tab:agent_added_value}) were environment setup (25 of 25 sessions), running code (23), identifying main scripts (20), and navigating the README and related files quickly (19). While not every reproduction required fixing errors, agents were perceived as adding value in such situations as well (e.g. "Debugging errors from running code as is" in 14 sessions). See \Cref{appendix:human_agent_example} for a few concrete examples where agents were able to resolve such operational blockers without human intervention.
    \item \textbf{The agent logged blockers more often than humans on the agent-only runs, but recovered more reliably.} Agents recorded at least one blocker category humans did not in 18 of the 19 papers where the agent was able to complete reproduction on its own. Across 34 paper-blocker category pairs, half fell in tooling or environment: headless-machine artifacts such as missing pdftotext or base R, JavaScript challenges while reading web pages, and slow package-manager progress being misread as hangs, which agents resolved without human intervention. On 39 occasions, the agent and the human encountered the same blocker category in a particular paper pair. Out of these, there were 11 instances where the agent fully recovered while the human only partially recovered or did not recover at all, six where the agent partially recovered while the human fully recovered (there were no instances where the agent completely failed to recover), and 22 where recovery (either full, partial, or none) was tied. The agent fully resolved missing or broken repository artifacts on four papers where humans could not. The agent fully or partially resolved all but 2 of the 114 individual blockers it encountered, while humans left 11 of 60 unresolved.
\end{enumerate}

Evaluators also answered other questions about each session, including which steps of the process had been performed solely by the agent and with what level of success (see \Cref{appendix: process_success}), and what kinds of struggles, if any, the agent had encountered in general (see \Cref{tab:agent_struggles}). Complementing the AI-assisted analysis of the full session logs reported in the fourth finding above (see \Cref{appendix:rubric}), evaluators also independently flagged session-level blockers they saw and whether each required human intervention (see \Cref{tab:blocker_summary,tab:blocker_examples}): 11 of 25 human-agent collaborative sessions involved at least one substantive blocker, and 10 of 30 blocker events required human intervention.

\subsection{Limitations}
\label{sec:limitations}

\textbf{Sample size limits generalizability.} The uplift study involves 20 papers and 5 participants, which limits the generalizability of our findings to broader populations of papers, fields, and researchers. While the estimated positive effect is statistically significant, the small sample size did not permit a serious investigation of potential heterogeneous effects. For example, agents might provide substantial uplift only in some of the fields included in the sample and not in others. 

\textbf{No ground truth results.} We did not have a verified ground truth for the paper reproduction attempts aside from the results in the paper. While this better reflects real-world computational reproducibility tasks and the primary goal of our study was to investigate process-level uplift, the lack of ground truth of code reproduction prevents us from assessing outcome correctness.

\textbf{Reproducers' backgrounds are non-representative.} The backgrounds of the reproducers may not reflect the broader population of researchers using agents for computational reproducibility. 

\textbf{The construct misses some benefits of manual reproduction.} The results of our randomized study show uplift of human-agent collaboration in completion time and recovery from blockers. However, these miss some benefits of manual code reproduction such as gaining an understanding of the codebase, data, or paper itself that may be important for certain types of reproduction tasks.

\textbf{Reproducers may be biased.} AI uplift study results are often vulnerable to participant biases due to the difficulty of fully blinding participants to AI treatment \citep{paskov2026rctshumanuplift}. In addition, since the reproducers in our randomized study are all also coauthors of this paper, demand effects could be possible. We tried to partially address this issue in the experiment plan by recording detailed terminal logs of both manual and human-agent collaborative sessions using Docent, which we make publicly available.

\textbf{Machine learning papers have a skewed distribution.} The machine learning papers in the study were drawn from award-winning conference papers, which is not representative of the broader literature. Award-winning papers may be better documented or more reproducible on average.

\textbf{Paper selection criteria are narrow and specific.} Our paper selection criteria include the requirement that the paper contains tables or figures with specific results that are suitable for defining clear success criteria for their reproduction, only Python or R tasks, and an estimated compute time of less than 45 minutes on the hardware used in our experiment. These are not representative of all computational reproducibility tasks (see \Cref{appendix:paper_selection} for the full paper criteria).

\section{Conclusion}
The dominant \emph{retire-and-replace} paradigm falls short of extracting robust information about agent performance beyond benchmark accuracy. Our premise is that this convention misses underlying dimensions of agent behavior that are crucial for informing deployment decisions. We propose essential steps towards measurement beyond accuracy saturation: investigating benchmark validity, evaluating agents in multiple dimensions (efficiency, reliability, and the relative importance of the model versus the scaffold), and measuring uplift from human-agent collaboration. Our aim is for these contributions to serve as a basis for moving past accuracy-centric evaluation.

\section{Acknowledgments}
This work was supported by Coefficient Giving, Schmidt Sciences, the Princeton AI Lab, the Princeton Language and Intelligence Initiative, and the Princeton Catalysis Initiative. We acknowledge compute credit from OpenAI. We thank Nicholas Carlini for identifying grading errors in CORE-Bench Hard and sharing a Claude Code scaffold that signaled accuracy saturation. We also thank Por Waiwitlikhit for contributing to the human-agent collaboration study. 

\bibliographystyle{plainnat} 
\bibliography{references_1,references_2}

\newpage
\appendix
\section{Technical appendices and supplementary material}
\subsection{Benchmark update details}
\label{appendix:all_changes}

We made the following changes to CORE-Bench Hard's grading script when grading agent responses in CORE-Bench v1.1 and CORE-Bench OOD:
\begin{enumerate}
    \item Expanded CORE-Bench Hard's original 95\% prediction interval to accept answers that lie within the default tolerances of \texttt{np.isclose} at the upper and lower bounds of the prediction interval.
    \item Expanded CORE-Bench Hard's original 95\% prediction interval to accept answers where agents reported unrounded results directly from computation when the ground truth was a rounded value.
    \item Checked if the ground truth answer was "True" or "False" as a \texttt{string}, and if the agent's answer was instead reported as a \texttt{boolean}. Converted the agent's answer to a \texttt{string} before grading (this only affected task \texttt{capsule-2242462}).
    \item Accepted multiple answers for the tasks in \Cref{tab:grading}.
\end{enumerate}

\subsection{Accuracy saturation of CORE-Bench v1.1 and CORE-Bench OOD}
\label{appendix:saturation_section}

We adopt metrics from \citet{akhtarWhenAIBenchmarks2026} that use the standard error of the difference in accuracy between the scores of top and $k$th agent to determine the similarity of accuracies on CORE-Bench v1.1 and CORE-Bench OOD. 

The standard error of the difference between the top and $k$th agent for $n$ benchmark tasks is:
\begin{gather*}
    \text{SE}_\Delta \approx \sqrt{\frac{s_1(1-s_1)}{n_{\text{eff}}} + \frac{s_k(1-s_k)}{n_{\text{eff}}}} \\
    \text{where } n_{\text{eff}} = n^\alpha, \alpha \in [0, 1], \text{ default } \alpha = 0.5 \\
    \text{and } s_1 \geq ... \geq s_k \text{ denotes the scores of the top } k \text{ agents}. 
\end{gather*}

The top $k$ agents are statistically indistinguishable in accuracy if:
\begin{gather*}
    s_1 - s_k \leq z \cdot \text{SE}_\Delta 
\end{gather*}

Using $\alpha=0.5$ and $z = 1.96$ for a 95\% confidence interval, we show that accuracies on both CORE-Bench v1.1 and CORE-Bench OOD for the top $k=5$ agents are statistically indistinguishable (see \Cref{table:saturation}). 

\begin{table}[t]
    \caption{\textbf{Updates to CORE-Bench Hard}. For each task, we compared the agent's accuracy to its computation and process correctness. We manually analyzed logs to identify the reason for the discrepancy between the original grade and the process or computation correctness.}
    \label{table:task_categorization}
    \smallskip
    \adjustbox{width=\textwidth}{
    \label{tab:regrading}
    \begin{tabular}{@{}p{2cm}p{2cm}p{2cm}p{6cm}p{3cm}p{3cm}p{3cm}@{}}
    \toprule
    \textbf{Original grade} & \textbf{Process correctness} & \textbf{Computation correctness} & \textbf{Possible explanation of discrepancy} & \textbf{Reason for discrepancy} & \textbf{Update}\\
    \midrule
    Correct & Correct & Correct & N/A & Neither & No changes \\
    \midrule
    Correct & Correct & Incorrect & The agent reproduced the paper correctly but ultimately used results from a pre-existing artifact for the answer. & Threat to construct validity & Remove task \\
    \midrule
    Correct & Incorrect & Correct & The agent reproduced only what was necessary to obtain the correct answer or wrote ad-hoc scripts to subvert needing to reproduce the entire paper's code. & Neither & No changes \\
    \midrule
    Correct & Incorrect & Incorrect & The agent was able to guess the answer or used results from a pre-existing artifact for the answer. & Threat to construct validity & Remove task \\
    \midrule
    Incorrect & Correct & Incorrect & The agent's process for reproducing the paper's code was correct, but ultimately made a computation error. & Agent error & No changes \\
    \midrule
    Incorrect & Incorrect & Correct & The agent incorrectly reported the answer. & Agent error & No changes \\
    \midrule
    Incorrect & Correct & Correct & The task prompt, ground truth, or grading contained errors. & Threat to construct validity & Edit the task or grading \\
    \midrule
    Incorrect & Incorrect & Incorrect & N/A & Agent error & No changes \\
    \midrule
    Incorrect & Unsolvable task & Unsolvable Task & The agent must access a dataset, library, or package that is not available. & Neither & Remove task \\
    \midrule
    \end{tabular}
    }
\end{table}

\begin{table}[t]
\label{table:validity_errors}
    \centering
    \caption{\textbf{Number of tasks affected by threats to construct validity in CORE-Bench Hard.} In total, we found \textbf{15} tasks with one or more task-level errors, and 20 tasks (four overlapping with the errors) where the answer can be trivially obtained from a pre-existing artifact. We removed 16 tasks and edited 15 tasks by either removing or editing only the affected task questions, editing the ground truth, or editing the grading script. These threats are difficult to surface prior to saturation, and we provide a few examples of how in \Cref{table:log_analysis}.}
    \smallskip
    \adjustbox{width=\textwidth}{
    \label{tab:num_tasks}
    \begin{tabular}{@{}p{3cm}p{7cm}p{7cm}p{3cm}@{}}
    \toprule
    \textbf{Error type} & \textbf{Explanation} & \textbf{Example} & \textbf{Num. tasks affected}\\
    \midrule
    Incorrect ground truth & The ground truth answer used for grading was incorrect. & The task requires the agent to report the highest y-axis value. The ground truth answer is the lowest y-axis value. & \centering\arraybackslash 1 \\
    \midrule
    Task question error or underspecification & The task question was unclear or contained an error. & The task requires the agent to report the best accuracy on a test dataset. It's unclear what test dataset the task is referring to. & \centering\arraybackslash 3 \\
    \midrule 
    Grading error & The 95\% prediction interval didn't capture floating point differences, rounding, or alternate task solutions. The task could have multiple correct answers. & The task requires the agent to report an accuracy. The accuracy value is present in two places in the results: a text output file where the value is not rounded, and a figure where the value is rounded. The agent reports the value from the text output file, but the ground truth answer is from the figure. &  \centering\arraybackslash 7  \\
    \midrule 
    Unsolvable task & The task relies on data, packages, or libraries that are not available. The results are non-deterministic. & The task requires the agent to download a dataset from a URL that is no longer live.  & \centering\arraybackslash 4 \\
    \midrule
    \end{tabular}
    }
\end{table}

\begin{table}[t]
    \centering
    \caption{\textbf{Examples of task-level threats to benchmark validity in CORE-Bench Hard and our initial version of CORE-Bench OOD.} These threats are difficult to surface with less capable agents. Prior to accuracy saturation, one of the most common failure cases on CORE-Bench Hard was agents unable to resolve version dependency conflicts \citep{siegel2024corebenchtmlr}. These agents were not progressing far enough into the reproduction pipeline to take shortcuts, report correct answers that were inaccurately marked as incorrect, or encounter environmental barriers. This made anticipating all task-level threats intractable before accuracy saturated.}
    \label{table:log_analysis}
    \smallskip
    \adjustbox{width=\textwidth}{
    \label{tab:grading}
    \begin{tabular}{@{}p{3cm}p{7cm}p{7cm}@{}}
    \toprule
    \textbf{Capsule ID} & \textbf{Task question} & \textbf{Error} \\
    \midrule
    \texttt{capsule-9670283} & \textit{From the final result plot, report the label for the blue line.} &  The agent is able to guess label from the color of the plot line using \texttt{matplotlib}'s default color order. \\
    \midrule
    \texttt{capsule-3262218} & \textit{Report the number of methods counter-arguments provided to defend the original study in light of the contradictory replication results.} & The agent could obtain the correct answer by running a trivial, ad-hoc command to count CSV rows where \texttt{methodsCounter == TRUE} without reproducing the paper's results. \\
    \midrule
    \texttt{capsule-4299879} & \textit{From the figure measuring bootstrapped predictive distribution of endline trust in police assuming mean regression at rate of mean regression among unexposed citizens, report the p-value from the Heard of Meetings plot.} & If the agent re-runs the bootstrap calculation in isolation without running the full end-to-end reproduction pipeline, the code will produce non-deterministic random samples because the seed is set at the beginning of the script. \\
    \midrule
    \texttt{capsule-5801588} & \textit{Report the label of the line from the plot measuring model evaluations at each iteration with the highest Model Evaluations at iteration 10.0.} & During benchmark construction, three code runs yielded the same task answer. However, when multiple agents were marked incorrect for this task with no apparent trajectory-level errors, we ran the script twice more and found the answer to be non-deterministic. \\
    \midrule
    \texttt{capsule-2675546} & \textit{From the ROC curve of UE \#74, report the true positive rate when the false positive rate is 0.4.} & The answer to this task question differs when the agent runs the script with Python 3.12 and newer libraries, versus the original paper's runs that use Python 3.6. \\
    \bottomrule
    \end{tabular}
    }
\end{table}

\begin{table}[t]
    \centering
    \caption{We updated the grading script for five capsules and six task questions to accept multiple answers. \texttt{capsule-2151475} is in CORE-Bench OOD. The rest are in CORE-Bench v1.1.}
    \smallskip
    \adjustbox{width=\textwidth}{
    \label{tab:grading}
    \begin{tabular}{@{}p{3cm}p{7cm}p{7cm}@{}}
    \toprule
    \textbf{Capsule ID} & \textbf{Task question} & \textbf{Reason for accepting multiple answers}\\
    \midrule
    \texttt{capsule-2816027} & \textit{For CTCF Signature Enrichment, report the name of the group with the highest median GSVA score.} & The group name in the capsule's figure label and the actual group name differ. We accept both. \\
    \midrule
    \texttt{capsule-3639589} & \textit{Report the color of the line with the highest maximum activation for target memory activation, DM.} & There are two plots in the results that show maximum activation for DM with different plot colors. We accept both. \\
    \midrule
    \multirow{2}{*}{\texttt{capsule-2151475}} & \textit{Report the name of the university ranked \#1 by impact factor.} & The ground truth is the abbreviation of the university name. We accept both the full university name and the abbreviation as it appears in the result figure. \\ \\
    \vspace{10pt}
    & \textit{Report the name of the journal with the highest 2011 impact factor from the analysis of 30 journals.} & The ground truth is the abbreviation of the journal name. We accept both the full journal name and the abbreviation as it appears in the result figure. \\
    \midrule
    \texttt{capsule-0152700} & \textit{Given the Kruskal-Wallis for Group 0-2 (Group 1 vs. Group 3), what is the p-value?} & The capsule results contain three deterministic p-values. We accept all three. \\
    \midrule
    \texttt{capsule-9477017} & \textit{Pearson correlation coefficients between the estimated proportions of different cell types were calculated, what is the highest Pearson correlation related to? Give the response in a list of strings.} & There are two possible highest correlation coefficients from the result plots. We accept cell types related to both, order-agnostic. \\
    \midrule
    \texttt{capsule-4252248} & \textit{Report the overall AUC from the PR curve generated with the CTRPv2 sensitivity dataset, tested against ATC annotations and drug-target information from CHEMBL.} & The AUC in the plot title is not rounded, but the AUC in the plot legend is rounded to three decimal places. We accept both. \\
    \bottomrule
    \end{tabular}
    }
\end{table}

\begin{figure}[t]
    \centering
    \begin{subfigure}[b]{\textwidth}
        \includegraphics[width=\linewidth]{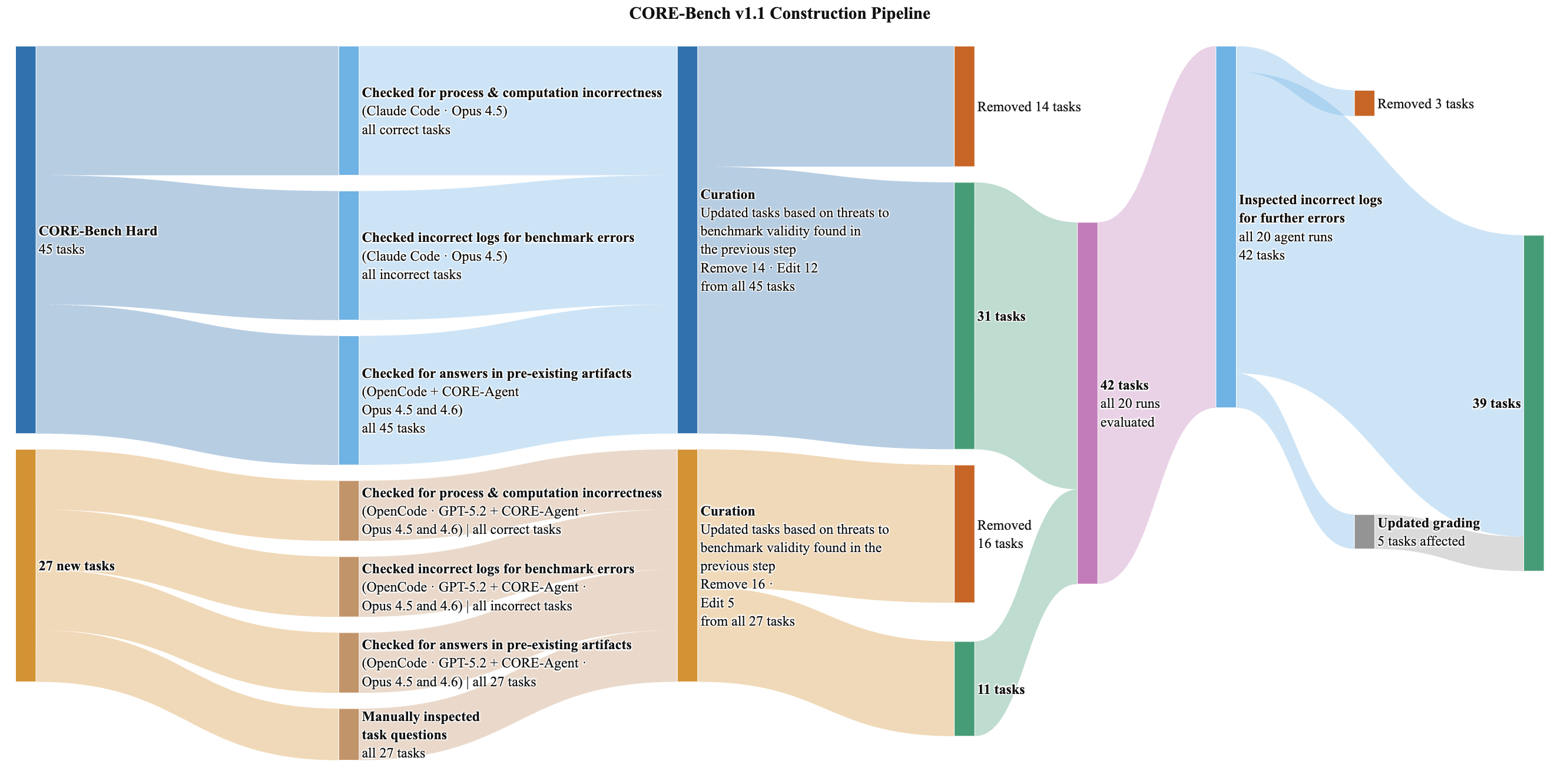}
        \caption{\textbf{CORE-Bench v1.1 construction pipeline.} We used automated and manual log analysis to identify threats to construct validity affecting the 45 CORE-Bench Hard tasks and 27 newly added tasks that informed updates and grading changes. These threats were difficult to surface with less capable agents that weren't progressing far enough past initial task solution steps to encounter errors or exploit shortcuts. The resulting benchmark, CORE-Bench v1.1, consists of 39 tasks that reflect validity improvements compared to the original dataset. We provide a summary of our rubrics (\Cref{table:docent_rubrics}) and other details on benchmark construction in \Cref{appendix:all_changes}}
        \label{figure:pipeline}
    \end{subfigure}

    \vspace{0.5em}

    \begin{subfigure}[b]{\textwidth}
        \includegraphics[width=\linewidth]{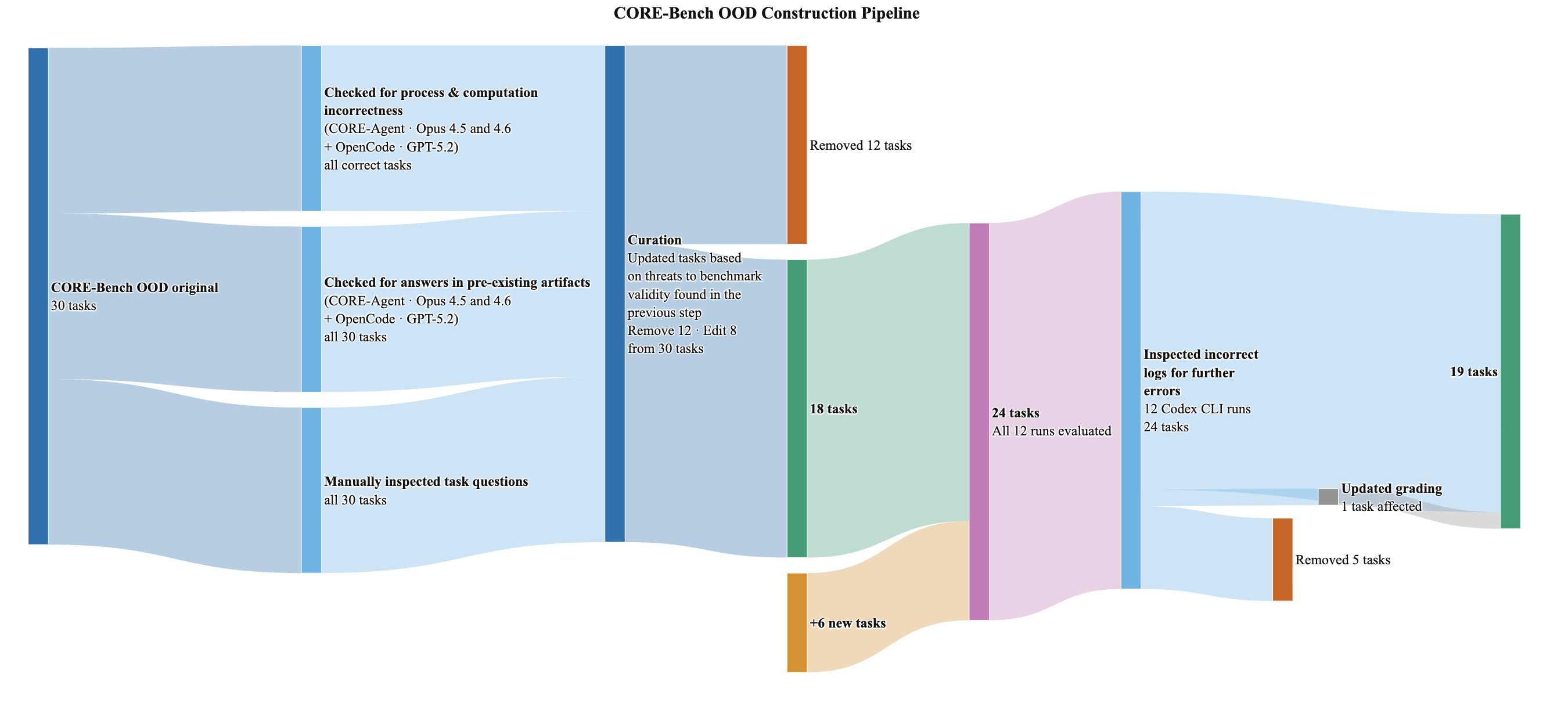}
        \caption{\textbf{CORE-Bench OOD construction pipeline.} We used a similar method of automated and manual log analysis as \Cref{figure:pipeline} to identify threats to benchmark validity affecting our original CORE-Bench OOD test set. The resulting benchmark, CORE-Bench OOD, has 19 tasks.}
        \label{fig:pipeline_ood}
    \end{subfigure}

    \caption{Construction pipelines for CORE-Bench v1.1 and CORE-Bench OOD.}
    \label{fig:pipelines}
\end{figure}

\begin{table}[t]
    \label{table:saturation_level}
    \scriptsize
    \centering
    \caption{\textbf{Saturation metrics.} We use the operationalization of saturation from \citet{akhtarWhenAIBenchmarks2026} to show that the top-five agents on both CORE-Bench v1.1 and CORE-Bench OOD have statistically indistinguishable accuracies.}
    \smallskip
    \adjustbox{width=\linewidth}{
    \label{table:saturation}
    \begin{tabular}{ccccccc}
    \toprule
    \textbf{\makecell{Benchmark}} & {$s_1$} & {$s_5$} & {$\Delta = s_1-s_5$} & {$z \cdot \text{SE}_\Delta$} & {$\Delta \leq z \cdot \text{SE}_\Delta$}  \\
    \midrule
    CORE-Bench v1.1 & 1 & 0.9744 & 0.0256 & 0.1240 & True \\
    \midrule
    CORE-Bench OOD & 1 & 0.8947 & 0.1053 & 0.2881 & True \\
    \bottomrule
    \end{tabular}
    }
\end{table}

\subsection{Benchmark implementation}
\label{appendix:core_implementation}
We run all agents on Azure virtual machines. The tasks requiring GPU are run on \texttt{Standard\_NC4as\_T4\_v3} and the remainder of the tasks are run on \texttt{Standard\_D4s\_v3}. All runs use the HAL evaluation harness \citep{kapoor2026holistic}. HAL provides a standard harness for reproducible agent evaluation and uses Weave for automated logging \citep{Wandb.ai2024Wandb}. All agents have full file system access and full web access.

For all Codex CLI, Claude Code, and OpenCode agents, we set per-task timeout to 45 minutes and max retries to 3. For CORE-Agent, we set per-task timeout to 5 hours, max steps to 200, and max retries to 1.

\subsubsection{Differences in results from Codex CLI versions}
\label{sec:codex_version}
We found that accuracy on CORE-Bench v1.1 with GPT-5.1 differed significantly based on Codex CLI version, with Codex CLI v0.122 obtaining an accuracy about 40\% higher than Codex CLI v0.130.0. Despite both versions using GPT-5.1, Codex CLI v0.130.0 had much shorter trajectories than Codex CLI v0.122: about two-thirds the total commands and one-fourth the output tokens.

In \Cref{sec:validity} and \Cref{sec:efficiency}, we report results using Codex CLI v0.122 for all Codex CLI runs. In \Cref{sec:reliability}, we report results using Codex CLI v0.130.0 for all models except GPT-5.1, where we use Codex CLI v0.122.

\subsection{Benchmark task breakdowns}
We provide a task breakdown of CORE-Bench v1.1 compared to CORE-Bench Hard in \Cref{table:task_breakdown}.
\begin{table}[t]
    \caption{Number of tasks by field and language in the test set of CORE-Bench Hard and CORE-Bench v1.1.}
    \label{table:task_breakdown}
    \begin{subtable}{0.55\textwidth}
    \centering
    \adjustbox{width=\textwidth}{
    \label{tab:field_task_comparison}
    \begin{tabular}{@{}p{3cm}p{2cm}p{2cm}p{2cm}p{1.75cm}@{}}
    & \makecell{Computer \\ Science} & \makecell{Social \\ Science} & \makecell{Medical \\ Science} & \makecell{Total} \\
    \toprule
    CORE-Bench Hard & \makecell{18} & \makecell{14} & \makecell{13} & \makecell{45}\\
    \midrule
    CORE-Bench v1.1 & \makecell{13} & \makecell{10} & \makecell{16} & \makecell{39} \\
    \midrule
    \end{tabular}
    }
    \caption{Task comparison by field}
    \end{subtable}
    \hfill
    \centering
    \begin{subtable}{0.42\textwidth}
    \adjustbox{width=\textwidth}{
    \label{tab:lang_task_comparison}
    \begin{tabular}{@{}p{3cm}p{1.75cm}p{1.75cm}p{1.75cm}@{}}
    & \makecell{Python} & \makecell{R} & \makecell{Total} \\
    \toprule
    CORE-Bench Hard & \makecell{22} & \makecell{23} & \makecell{45} \\
    \midrule
    CORE-Bench v1.1 & \makecell{18} & \makecell{21} & \makecell{39}\\
    \midrule
    \end{tabular}
    }
    \caption{Task comparison by language}
    \end{subtable}
\end{table}

\subsection{Randomized study details}
\label{sec:uplift_appendix}
We provide additional details on methodology and implementation of the uplift study.
\subsubsection{Paper Selection Criteria}
\label{appendix:paper_selection}
A paper was included only if all of the following criteria were met:
\begin{enumerate}
    \item \textbf{From sources:}
    \begin{enumerate}[label=\alph*.]
        \item For ML papers: Won a paper award at one of these conferences, AAAI, ACL, CVPR, ECCV, EMNLP, ICCV, ICLR, ICML, IJCAI-JAIR, NeurIPS, and 3DV from 2011--2025 (sourced from \url{https://github.com/clemense/ai-bestpapers})
        \item For non-ML papers: evaluated in the I4R reproducibility study (and found to be ``evaluable'' there, e.g.\ data available)
    \end{enumerate}
    \item \textbf{GitHub (or other) repository exists with code}
    \item \textbf{Can run on single GPU or CPU} (no hosted models)
    \item \textbf{More specifically:} Reproduction of targets we selected from the paper (see below) looks likely to run in our setup (A40 48GB VRAM, disk space: 40GB+40GB - see evaluator instructions for details)
    \item \textbf{Data available (link). (where applicable)}
    \begin{enumerate}[label=\alph*.]
        \item ``Available'' meaning for direct download without registration or such
        \item For ML papers, this might include pre-existing benchmarks (e.g.\ for \url{https://arxiv.org/pdf/2312.12337} this could be the RealEstate10k dataset from an earlier paper)
    \end{enumerate}
    \item \textbf{Pretrained weights available (link)} (where applicable). Notes:
    \begin{enumerate}[label=\alph*.]
        \item ``Available'' meaning for direct download without registration or such
    \end{enumerate}
    \item \textbf{Uses Python or R}
    \item \textbf{Clear success criteria} (specific tables/figures)
    \item \textbf{Not previously seen} by the evaluator (defined as having read at most the abstract)
    \item \textbf{Compute time limit}: running the code / inference necessary for the reproduction is anticipated to take less than 45 minutes on our hardware. Notes:
    \begin{enumerate}[label=\alph*.]
        \item ``Compute time'' refers to the cumulative duration of the agent and/or human evaluator having to wait for VM to complete compute tasks.
        \item This represents the compute reproduction time for all replication targets together.
        \item Does not include Run 2 \& Run 3 for non-deterministic outcomes if floating point tolerance criterion (see evaluator instructions) is not met.
        \item Does not include wait times for data or model downloads.
        \item Does not include the time the agent spends reasoning or using other tools.
        \item Estimates are OK (e.g.\ concluding that this criterion is not met after a progress bar shows 10\% completed after 10 minutes).
        \item Does not include the environment set up and dependencies
    \end{enumerate}
\end{enumerate}

\subsubsection{Uplift study implementation}
\label{appendix:uplift_implementation}

For the uplift study, both human-only (manual) and human-agent reproduction attempts are run inside standardized Docker environments to ensure consistency across participants and conditions. ML papers (from AI conferences) are run on cloud GPU instances using A40 GPUs on RunPod with a dedicated ML Docker image, while non-ML papers (from the I4R source) are run using a separate non-ML Docker image; both templates are configured with 40\,GB container disk (plus 40\,GB volume for the ML template). Using a uniform GPU and VM configuration ensures that runs are comparable in compute resources.

In the human--agent (AI-allowed) condition, participants use Codex CLI with the GPT-5.4 model at the ``extra high'' reasoning setting. Codex is launched inside the Docker container using the harness available at \url{https://github.com/ab-shetty/agent-reproducibility}, which automates session logging and uploads traces to Docent for later inspection. Each participant provides the paper PDF, the default reproduction prompt (Appendix~\ref{appendix:default_prompt}), and the replication target.

In the manual (AI-disallowed) condition, participants may use only traditional web resources such as documentation, forums and StackOverflow; no generative AI tools (e.g., ChatGPT, GitHub Copilot, Claude) and no AI-generated search summaries (e.g., Google AI Overviews, Bing Copilot) are permitted. To suppress AI Overviews, participants append the \texttt{-ai} flag to every Google query. Non-generative IDE autocomplete is allowed.

To preserve independence across attempts, a fresh pod is launched for each reproduction, or the current pod is fully reset before reuse. The maximum time limit for a single reproduction attempt is 3 hours, after which the attempt is recorded as unsuccessful. Following each run, participants complete a structured post-run questionnaire (Appendix~\ref{appendix: Questionnaire}) capturing their experience, blockers encountered and self-reported confidence in the result.

\subsubsection{Default Prompt}
\label{appendix:default_prompt}
The following prompt is provided to participants (and, in the agent condition, to the agent) at the start of each reproduction attempt. The placeholders \texttt{[PAPER\_NAME]}, \texttt{[REPO\_URL]}, \texttt{``replication target''}, and \texttt{\{replication target\}} are filled in per task.

\begin{quote}
\setlength{\parskip}{1em}
Below is a result (``replication target'') selected from the research paper present in this directory, titled ``[PAPER\_NAME]''. Reproduce this replication target exclusively by running the paper's code. All we care about is getting there through genuine reproduction.

Obtain the code here: [REPO\_URL]

Read the README (if present). Set up the environment, install dependencies, download any required data, then run the code to reproduce the following result (``replication target'') reported in the paper:

\{replication target\}

Rules:
\begin{itemize}[label={-}, itemsep=1em]
    \item Do not modify any script's scientific logic. Limit changes to environment compatibility only (e.g.\ dependency versions, paths, deprecated APIs, config variables, runtime arguments/variables such as model type or dataset).
    \item If stuck after 2-3 attempts on the same error, stop and tell me what's wrong so we can figure it out together.
    \item Save all generated outputs and report back the results. For numeric values, report the exact value of the output - do not round or truncate
    \item If the reproduction value of the replication target is not within the floating point tolerance of 1e5 * sys.float\_info.epsilon of the paper's reported value after rounding the reproduction value to the same number of decimal places, then run 2 more times and determine if the reproduction value falls within the 95\% prediction interval using the \_compute\_prediction\_intervals function below.
\end{itemize}
\end{quote}

\begin{quote}
\begin{lstlisting}[
    language=Python,
    basicstyle=\ttfamily\small,
    keywordstyle=\ttfamily\bfseries,
    stringstyle=\ttfamily,
    commentstyle=\ttfamily\itshape,
    columns=fullflexible,
    keepspaces=true,
    showstringspaces=false,
    breaklines=true,
    label={lst:prediction_intervals}
]
def _compute_prediction_intervals(
    reproduction_values: list[dict],
    numeric_keys: list[str]
) -> dict[str, dict]:
    """Compute 95%
    intervals = {}
    sample_size = len(reproduction_values)
    if sample_size < 2:
        for key in numeric_keys:
            value = reproduction_values[0].get(key, 0)
            intervals[key] = {"lower": value, "upper": value, "mean": value}
        return intervals
    t_value = t.ppf(0.975, sample_size - 1)
    for key in numeric_keys:
        values = [rv.get(key, 0) for rv in reproduction_values]
        mean = np.mean(values)
        std = np.std(values, ddof=1)
        margin = t_value * std * math.sqrt(1 + 1 / sample_size)
        intervals[key] = {
            "lower": mean - margin,
            "upper": mean + margin,
            "mean": mean,
        }
\end{lstlisting}
\end{quote}

\subsubsection{Papers selected for reproduction}
\label{app:PaperTargets}

{
\small
\setlength\LTleft{0pt}
\setlength\LTright{0pt}
\begin{longtable}{@{}p{4cm} p{1.3cm} p{4.5cm} p{0.6cm} p{1.5cm}@{}}
\caption{\textbf{Papers selected for reproduction, with field and reproduction target.} Targets were chosen by selectors by picking a specified value from the published paper. The final column records the observed outcome from our study: ``Matched'' indicates that at least one reproduction attempt achieved the target metric within the specified tolerance; ``Result, no match'' indicates that at least one attempt produced a result but none matched within tolerance; and ``No results produced'' indicates that no attempt produced a usable result.}
\label{tab:reproduction-papers}\\
\toprule
\thead[l]{\textbf{Paper}} & \thead[l]{\textbf{Field}} & \thead[l]{\textbf{Reproduction Target}} & \thead[l]{\textbf{Paper}\\\textbf{Code}} & \thead[l]{\textbf{Observed}\\\textbf{outcome}} \\
\midrule
\endfirsthead

\toprule
\thead[l]{\textbf{Paper}} & \thead[l]{\textbf{Field}} & \thead[l]{\textbf{Reproduction Target}} & \thead[l]{\textbf{Paper}\\\textbf{Code}} & \thead[l]{\textbf{Observed}\\\textbf{outcome}}\\
\midrule
\endhead

\midrule
\multicolumn{5}{r@{}}{\textit{Continued on next page}} \\
\endfoot

\bottomrule
\endlastfoot

Obfuscated gradients give a false sense of security: Circumventing defenses to adversarial examples\cite{athalye2018obfuscated} & Machine Learning & Accuracy under the defense from Buckman et al.\ (2018) on CIFAR (Table 1): 0\% & \href{https://github.com/anishathalye/obfuscated-gradients}{Code} & No results produced\\[0.5em]

Latxa: An open language model and evaluation suite for basque\cite{etxaniz2024latxa} & Machine Learning & Performance of Latxa 7B on EusProf (Table 1): 30.26 & \href{https://github.com/hitz-zentroa/latxa}{Code} & Result, no match\\[0.5em]

Beyond accuracy: Behavioral testing of NLP models with CheckList\cite{ribeiro2020checklist} & Machine Learning & Failure rate of BERT-base on Sentiment Analysis ``Negated neutral should still be neutral'' MFT (Table 1): 98.4\% & \href{https://github.com/marcotcr/checklist}{Code} & Matched\\[0.5em]

Semisupervised neural proto-language reconstruction\cite{meloni2024semisupervised} & Machine Learning & TED of Transformer DPD-$\Pi$M-BST on 10\% labeled WikiHan, averaged across all runs in four groups (Table 2): 1.0075 & \href{https://github.com/cmu-llab/dpd}{Code} & Matched \\[0.5em]

Improving evaluation of machine translation quality estimation\cite{graham2015improving} & Machine Learning & Williams test outcome for HTER prediction in EN$\to$ES WMT-14 Task 1.2: significant increase in Pearson correlation for HTER-DCU-rtm-svr over HTER-DCU-rtm-tree ($p < 0.05$) & \href{https://github.com/ygraham/mt-qe-eval}{Code} & Matched \\[0.5em]

Reliable conflictive multi view learning\cite{si2024reliable} & Machine Learning & Conflictive test-set accuracy of ECML on Scene15 (Table 3): $56.97 \pm 0.52\%$ & \href{https://github.com/jiajunsi/RCML}{Code} & Matched \\[0.5em]

Fantastically ordered prompts and where to find them: Overcoming few-shot prompt order  sensitivity\cite{lu2022fantastically} & Machine Learning & Performance of GPT-2 0.1B GlobalE on Template 1 (Table 3): 63.8 & \href{https://github.com/yaolu/ordered-prompt}{Code} & Matched\\[0.5em]

Informer: Beyond efficient transformer for long sequence time-series forecasting\cite{zhou2021informer} & Machine Learning & MSE of Informer on ETTh1 with 24 counts (Table 2): 0.577 & \href{https://github.com/zhouhaoyi/Informer2020}{Code} & Matched \\[0.5em]

DropMessage: Unifying random dropping for graph neural networks\cite{fang2023dropmessage} & Machine Learning & Accuracy of GCN-DropMessage on PubMed (Table 2): 79.20 & \href{https://github.com/zjunet/DropMessage}{Code} & Matched \\[0.5em]

MultiWOZ — a large-scale multi-domain wizard-of-oz dataset for task-oriented dialogue modelling\cite{budzianowski2018multiwoz} & Machine Learning & Number of dialogues in the MultiWOZ training split (Table 1): 8,438 & \href{https://github.com/budzianowski/multiwoz}{Code} & Matched\\[0.5em]

\midrule

Talking shops: The effects of caucus discussion on policy coalitions\cite{kalla2022talking} & Social Science & Deliberation effect on cosponsorship for attended meetings, non-sponsor's party (Table 4): 5.9 pp & \href{https://dataverse.harvard.edu/dataset.xhtml?persistentId=doi:10.7910/DVN/S3M5AX}{Code} & Matched\\[0.5em]

Decentralization can increase cooperation among public officials\cite{grossman2021decentralization} & Social Science & Coefficient for Decentralized in Weighted Poisson Full Model for Strong Ties (Table 3): 1.07 & \href{https://dataverse.harvard.edu/dataset.xhtml?persistentId=doi:10.7910/DVN/ZLHYSZ}{Code} & Matched\\[0.5em]

Changing tides: Public attitudes on climate migration\cite{hoffmann2022changing} & Social Science & AMCE for flooding vs.\ economic opportunity as migration reason, German sample (Table 2): 0.086 & \href{https://dataverse.harvard.edu/dataset.xhtml?persistentId=doi:10.7910/DVN/FDML2N}{Code} & Matched\\[0.5em]

Multiracial identity and political  preferences\cite{davenport2021multiracial} & Social Science & Whether White-Blacks are more conservative or more liberal than Blacks on police perceptions (Figure 1): more conservative & \href{https://dataverse.harvard.edu/dataset.xhtml?persistentId=doi:10.7910/DVN/BLVJJH}{Code} & Matched \\[0.5em]

Entertaining beliefs in economic mobility\cite{frank2022entertaining} & Social Science & Coefficient of Rags-to-Riches TV Treatment on belief in economic mobility, lab-in-the-field sample (Table 1, Col.\ 5): 0.068 & \href{https://dataverse.harvard.edu/dataset.xhtml?persistentId=doi:10.7910/DVN/FVRZYU}{Code} & Matched\\[0.5em]

Antinormative messaging, group cues, and the nuclear ban treaty\cite{mehrl2022antinormative} & Social Science & Treatment effect of Institution Cue on support for TPNW (Appendix Table H1, Model 2): $-19.2$ pp & \href{https://dataverse.harvard.edu/dataset.xhtml?persistentId=doi:10.7910/DVN/GLT4FX}{Code} & Matched\\[0.5em]

Policy deliberation and voter persuasion: Experimental evidence from an election in the Philippines\cite{arias2019policy} & Social Science & ITT of Vote (Akbayan) (Table 1): 1.955 & \href{https://dataverse.harvard.edu/dataset.xhtml?persistentId=doi:10.7910/DVN/S3HACJ}{Code} & Matched\\[0.5em]

Can’t we all just get along? how women MPs can ameliorate affective polarization in western publics\cite{weeks2022cant} & Social Science & Coefficient for out-party proportion of women MPs ($t-1$) among women partisans (Table 1): 2.1 & \href{https://dataverse.harvard.edu/dataset.xhtml?persistentId=doi:10.7910/DVN/AHQRVR}{Code} & Matched\\[0.5em]

 Indecent disclosures: Anticorruption reforms and political selection\cite{gulzar2021indecent} & Social Science & Treatment group $\times$ Second period election coefficient (Table 1): $-0.057$ (0.015) & \href{https://dataverse.harvard.edu/dataset.xhtml?persistentId=doi:10.7910/DVN/KDUMRM}{Code} & Matched \\[0.5em]

Yellow vests, pessimistic beliefs, and carbon tax aversion\cite{douenne2022yellow} & Social Science & OLS coefficient for Yellow Vests: supports (Table 2): $-0.108$ (0.026) & \href{https://github.com/thomasdouenne/yellow_vests_aej_ep}{Code} & Matched \\[0.5em]

\end{longtable}
}

\subsubsection{Instructions for evaluators (for manual and agent-based runs)}
\label{appendix:instructions}

\begin{itemize}[leftmargin=1.5em]
\item We're using Codex (with OpenAI credits)
\item Reproduction should be run in one of two Docker images (for ML/non-ML papers): (when using Runpod, this is already integrated into the template, see below)
\item Starting codex in the docker image (choose \texttt{gpt-5.4} with extra high thinking)
\item Reproductions of ML papers should be run on a cloud GPU environment like Lambda (which we already have credits for) or RunPod (AWS and Google Cloud also work). Currently we have planned to use \textbf{A40s on RunPod}. Using the same kind of GPU and VM ensures that runs are comparable in that respect.
\item Runpod
\begin{itemize}[leftmargin=1.5em]
\item Create a Runpod account and configure SSH
\item Use ML template/Docker image for the AI conference papers \url{https://console.runpod.io/deploy?type=GPU&gpu=A40&count=1&template=jo8klw71d0}. Under ``Select an instance'', pick the A40 GPU option. The template sets disk space at 40GB Container + 40GB Volume.
\item Use Non-ML template/Docker image for the I4R papers \url{https://console.runpod.io/deploy?type=GPU&gpu=A40&count=1&template=071bthitdg}. Under ``Select an instance'', pick the A40 GPU option.
\par The template sets disk space at 40GB Container.
\item Keep the pod running to prevent data loss through the course of reproduction (Codex logs, etc.\ not mounted in runpod \texttt{/workspace} directory)
\item If you have to download a file from your Runpod instance for inspection etc.:
\par Step 1: Install \texttt{runpodctl} (on your local machine)
\begin{verbatim}
mkdir -p ~/.local/bin && \
curl -sL https://github.com/runpod/runpodctl/\
releases/latest/download/\
runpodctl-linux-amd64.tar.gz \
| tar xz -C ~/.local/bin
export PATH="$HOME/.local/bin:$PATH"
runpodctl version
\end{verbatim}
\par Step 2: On your runpod instance run
\begin{verbatim}
runpodctl send ~/test_image.png
# outputs something like:
# Code is: 3476-quiet-telex-premium-9
# On the other computer run

# runpodctl receive 3476-quiet-telex-premium-9
\end{verbatim}
\par Step 3: On your local computer run
\begin{verbatim}
runpodctl receive 3476-quiet-telex-premium-9
\end{verbatim}
\end{itemize}
\item The Docker image will launch an internal script that automates logging and will ask for your Docent API key to upload traces/logs to Docent, including run metadata, once the user enters \texttt{finish-session}. Details in \texttt{README.md}.
\begin{itemize}[leftmargin=1.5em]
\item Will be using the final Docent collection for final runs. For the pilot we will be using pilot.
\end{itemize}
\item Upload a PDF of the paper (for Runpod, run the \texttt{runpodctl} tool on your local machine: \texttt{runpodctl send paper.pdf})
\item Launch Codex in the terminal, per the hint provided by the script.
\item Use \texttt{/model} to switch to \texttt{gpt-5.4} with extra high thinking
\item Keep the pod (VM) running during the reproduction attempt
\item After the reproduction attempt, use the \texttt{finish-session} command in the terminal to log the (raw) duration and upload the session log to Docent (optional when doing testing/pilot runs). Be sure to exit from your virtual environments before running the command.
\item After the run, fill out the questionnaire in Google Forms [see \Cref{appendix: Questionnaire}]
\item Launch a new instance for each reproduction, OR completely reset the current one (make sure it will be returned to the same state as after a new deployment, so as not to affect consistency)
\begin{itemize}[leftmargin=1.5em]
\item On Runpod, resetting can be achieved by
\begin{itemize}[leftmargin=1.5em]
\item wiping the workspace (i.e.\ delete all files and folders, using a command like \texttt{find /workspace -mindepth 1 -maxdepth 1 -exec rm -rf - \{\} +}), and then
\item using ``edit a pod'' (\url{https://docs.runpod.io/pods/manage-pods#update-a-pod} and saving the configuration again with no changes
\end{itemize}
\end{itemize}
\item Maximum time limit for a single reproduction attempt (referring to ``duration'' as defined below): 3 hours
\end{itemize}

\paragraph{Manual Condition (AI-Disallowed)}

\begin{itemize}[leftmargin=1.5em]
\item No generative AI tools used (e.g., GitHub Copilot, ChatGPT)
\item No AI-generated search summaries used (e.g., Google AI Overviews, Bing Copilot)
\item Only traditional web search (links, docs, StackOverflow) used
\begin{itemize}[leftmargin=1.5em]
\item In Google searches, append the ``-ai'' flag to every search to suppress automatic AI-generated results
\end{itemize}
\item No AI-based code generation or debugging assistance
\item All reproduction tasks must be executed within a standardized Docker environment.
\end{itemize}

\noindent Allowed: non-generative IDE autocomplete, documentation, forums

\subsubsection{Blockers Review Rubric}
\label{appendix:rubric}
We used the following rubric to select blockers from our logs of the human-agent collaboration runs, with the assistance of Codex using GPT-5.4. We define operational blockers as any concrete obstacle that delayed progress, forced a workaround or required debugging or rerouting. Agents encountered operational blockers 122 times across 25 runs (4.88 per run), identified using Codex-assisted log analysis. 74 arose during setup, 40 during execution and only 8 during result extraction or reporting. In practice, we saw that the agent's contribution was usually to repair the local reproduction path rather than simply launch a clean released pipeline.    

\begin{verbatim}
# Blocker Review Rubric

Each AI run should be reviewed independently from its exported transcript JSON.

Goal: extract a comprehensive but disciplined list of blockers the AI agent faced.

Definition of blocker:
- Any concrete obstacle that delayed progress, forced a workaround, caused a
  failed attempt, or required debugging/rerouting.
- Include both root-cause blockers and shorter-lived operational blockers if
  they materially interrupted the run.
- Do not include ordinary progress steps that were not obstacles.
- Do not include purely hypothetical risks unless they became an actual
  impediment in the transcript.

Granularity rule:
- Split distinct obstacles into separate blocker entries when they required
  different fixes or occurred in different phases.
- If multiple symptoms clearly stem from one issue, keep them in one blocker
  entry and describe the symptoms in the evidence.

Required output fields per run:
- collection_name
- actual_collection_name
- researcher
- paper
- agent_run_id
- agent_run_name
- model
- overall_outcome
- notes
- blockers

Required fields per blocker:
- label
- category
- phase
- resolved
- description
- evidence

Allowed category values:
- environment
- dependency
- repo_artifact
- path_config
- data_input
- runtime
- tooling

Allowed phase values:
- setup
- execution
- postprocess

Allowed resolved values:
- yes
- no
- partial

Evidence rule:
- Cite concrete transcript evidence in plain text, ideally with message indices
  or an explicit quoted/paraphrased action/result.
- Keep evidence concise.
\end{verbatim}

\subsubsection{Randomized study design}
\label{appendix:RCTdesign}
The randomized study aims to estimate the uplift effect of human-agent collaboration on the task of computationally reproducing a result (replication) target from a given paper. It was designed assuming that both the papers (replication targets) and the researchers (evaluators) carrying out the reproduction task may have unobserved characteristics that affect task duration. This motivates the use of a blocked randomization assignment with blocking on both researchers and papers, where the sampling of paper-evaluator pairs (among all possible combinations) and their random assignment to either the treatment (human-agent collaboration) or control (Manual) condition is restricted by the following balancing requirements:
\begin{itemize}
\item{Each of the 20 papers was assigned to either 2 or 3 of the 5 evaluators, and to each condition (manual or human-agent collaborative) at least once.}
\item{Each of the evaluators was assigned 10 papers (5 from each source), and to each condition (manual or human-agent collaborative) 5 times.}
\end{itemize}
To mitigate learning effects, evaluators were instructed to carry out the tasks in a specified randomized order.

The same five authors who acted as evaluators also carried out the selection of papers from the aforementioned two sources. This task included vetting of the predefined selection criteria (such as the availability of code and data for the paper, or that the reproduction should be feasible on the hardware used in the experiment), and selection of one specific result to replicate from each paper.
This process was designed to ensure blinding (i.e. as an additional constraint on the randomized assignment, no team member was assigned a paper as evaluator that they had encountered during the paper selection process). To achieve a degree of representativeness with respect to the given source and criteria, selectors were assigned a randomly selected and randomly ordered slice from each dataset, to assess for eligibility in the given order.

\subsubsection{Fixed effects model to estimate uplift}
\label{appendix:fixed_effects}
To estimate the uplift, we use linear regression with log task duration as the outcome variable (see \Cref{appendix: Questionnaire} for its exact definition as part of the evaluator questionnaire).
Our aforementioned assumption (that task duration might be affected by unobserved characteristics of both papers and evaluators) also motivates the use of a fixed effects model here, with fixed effects for both researchers and papers:

\begin{align}
\log(\mathrm{duration}_{i}) =
\alpha
+ \beta \, \mathrm{AI}_{i}
+ \gamma_p
+ \delta_r
+ \varepsilon_i,
\end{align}

Here $i$ indexes the individual replication session (identified with a paper-evaluator pair), $p$ denotes the paper being reproduced in session $i$, and $r$ the evaluator conducting the session. The terms $\gamma_p$ and $\delta_r$ are paper and evaluator fixed effects, respectively, and $\beta$ is the estimated difference between log task duration in the Manual condition relative to the Human-agent collaborative condition. We conceive uplift as a speed change here, and speed is reciprocal to duration; so, for simplicity we estimate the reciprocal factor for duration - from Human-agent collaborative to Manual instead of vice versa.

We use CR2 standard errors clustered by researcher. CR2 standard errors are designed for use with small-sample fixed effect models, and represent a conservative choice relative to conventional clustered standard errors.\citep{Pustejovsky02102018} They are the recommended small-sample correction in R's clubSandwich package.\citep{clubSandwich}

The model's coefficient estimate for the Manual condition is 0.7485, with a CR2 standard error of 0.0919, Satterthwaite degrees of freedom of 3.7, and p-value 0.00176. The point estimate implies that Manual sessions last about 2.11 times as long as human-agent collaborative sessions.

In the experiment, session duration was capped at 180 minutes, i.e. the outcome variable is right-censored. We did not attempt to account for this in the model. Because only manual runs hit this limit in our experiment, our uplift estimate is conservative in that regard.

\subsection{Questionnaire for human-agent reproducibility evaluation}
\label{appendix: Questionnaire}
\begin{verbatim}
=========================================================
Part I: Metadata and Environment Setup (Questions 1--48)
=========================================================

1  Paper Title
2  Link to the paper's code (GitHub or other)
3  Domain: {AI Conferences, I4R}
4  Email
5  Human Researcher [your name]

6  Hardware (Select "Other" only if you used an environment
   other than A40 on RunPod.):
     {GPU: A40 48GB VRAM CPU: Intel(R) Xeon(R) Gold 6342 CPU
      @ 2.80GH, Other}

7  OS (Select "Other" only if you used an environment other
   than A40 on RunPod.):
     {Ubuntu 22.04, Other}

8  Execution Environment (Select "Other" only if you performed
   a non-standardized step.):
     {Docker instance with pre-installed libraries, Other}

9  Date (PST) {mm, dd, yyyy}
10 Start Time (PST)
11 End Time (PST)
12 Link to Docent log of this session
13 Condition {Manual, AI-assisted}
14 Agent version: {gpt-5.4-codex with extra high thinking,
   Other}

15 Step 1.1: Start the instance and Docker image (Human)
   -- Outcome {Success, Failure, Partial Success}

16 Step 1.1: Start the instance and Docker image (Human)
   -- Notes

17 Step 1.2: Start Agent with logging and prompt replication
   target task (using the generic default prompt) (Human)
   -- Outcome
     {Success, Failure, Partial Success}

18 Step 1.2: Start Agent with logging and prompt replication
   target task (using the generic default prompt) (Human)
   -- Notes

19 Step 1.3: Obtain the paper's code (e.g. clone repo)
   -- Who did it {Human, Agent, Both}

20 Step 1.3: Obtain the paper's code (e.g. clone repo)
   -- Outcome {Success, Failure, Partial Success}

21 Step 1.3: Obtain the paper's code (e.g. clone repo)
   -- Notes

22 Step 1.4: Read README
   -- Who did it {Human, Agent, Both}

23 Step 1.4: Read README
   -- Outcome {Success, Failure, Partial Success}

24 Step 1.4: Read README -- Notes

25 Step 1.5: Create environment (e.g. using conda/venv)
   -- Who did it {Human, Agent, Both}

26 Step 1.5: Create environment (e.g. using conda/venv)
   -- Outcome {Success, Failure, Partial Success}

27 Step 1.5: Create environment (e.g. using conda/venv)
   -- Notes

28 Step 1.6: Install dependencies
   -- Who did it {Human, Agent, Both}

29 Step 1.6: Install dependencies
   -- Outcome {Success, Failure, Partial Success}

30 Step 1.6: Install dependencies -- Notes

31 Step 1.7: Download/prepare data
   -- Who did it {Human, Agent, Both}

32 Step 1.7: Download/prepare data
   -- Outcome {Success, Failure, Partial Success}

33 Step 1.7: Download/prepare data -- Notes

34 Step 1.8: Verify setup (import test, etc.)
   -- Who did it {Human, Agent, Both}

35 Step 1.8: Verify setup (import test, etc.)
   -- Outcome {Success, Failure, Partial Success}

36 Step 1.8: Verify setup (import test, etc.) -- Notes

37 Phase 1 Blocker 1: What was the blocker
38 Phase 1 Blocker 1: Who got stuck {Human, Agent, Both}
39 Phase 1 Blocker 1: What was the resolution
40 Phase 1 Blocker 1: Intervention needed? {Yes, No}

41 Phase 1 Blocker 2: What was the blocker
42 Phase 1 Blocker 2: Who got stuck {Human, Agent, Both}
43 Phase 1 Blocker 2: What was the resolution
44 Phase 1 Blocker 2: Intervention needed? {Yes, No}

45 Phase 1 Blocker 3: What was the blocker
46 Phase 1 Blocker 3: Who got stuck {Human, Agent, Both}
47 Phase 1 Blocker 3: What was the resolution
48 Phase 1 Blocker 3: Intervention needed? {Yes, No}

=========================================================
Part II: Reproduction Execution and Runtime Debugging
(Questions 49--75)
=========================================================

49 Phase 1 Step 2.1: Identify entry point / main script
   -- Who did it {Human, Agent, Both}

50 Phase 1 Step 2.1: Identify entry point / main script
   -- Outcome {Success, Failure, Partial Success}

51 Phase 1 Step 2.1: Identify entry point / main script
   -- Notes

52 Phase 1 Step 2.2: Understand required run parameters
   -- Who did it {Human, Agent, Both}

53 Phase 1 Step 2.2: Understand required run parameters
   -- Outcome {Success, Failure, Partial Success}

54 Phase 1 Step 2.2: Understand required run parameters
   -- Notes

55 Phase 1 Step 2.3: Run code
   -- Who did it {Human, Agent, Both}

56 Phase 1 Step 2.3: Run code
   -- Outcome {Success, Failure, Partial Success}

57 Phase 1 Step 2.3: Run code -- Notes

58 Phase 1 Step 2.4: Monitor / debug runtime errors
   -- Who did it {Human, Agent, Both}

59 Phase 1 Step 2.4: Monitor / debug runtime errors
   -- Outcome {Success, Failure, Partial Success}

60 Phase 1 Step 2.4: Monitor / debug runtime errors
   -- Notes

61 Phase 1 Step 2.5: Locate output files
   -- Who did it {Human, Agent, Both}

62 Phase 1 Step 2.5: Locate output files
   -- Outcome {Success, Failure, Partial Success}

63 Phase 1 Step 2.5: Locate output files -- Notes

64 Phase 2 Blocker 1: What was the blocker
65 Phase 2 Blocker 1: Who got stuck {Human, Agent, Both}
66 Phase 2 Blocker 1: What was the resolution
67 Phase 2 Blocker 1: Intervention needed? {Yes, No}

68 Phase 2 Blocker 2: What was the blocker
69 Phase 2 Blocker 2: Who got stuck {Human, Agent, Both}
70 Phase 2 Blocker 2: What was the resolution
71 Phase 2 Blocker 2: Intervention needed? {Yes, No}

72 Phase 2 Blocker 3: What was the blocker
73 Phase 2 Blocker 3: Who got stuck {Human, Agent, Both}
74 Phase 2 Blocker 3: What was the resolution
75 Phase 2 Blocker 3: Intervention needed? {Yes, No}

=========================================================
Part III: Result Evaluation and Blockers
(Questions 76--96)
=========================================================

76 Phase 3 Step 3.1: Parse/extract our results
   -- Who did it {Human, Agent, Both,
   N/A -- no results produced}

77 Phase 3 Step 3.1: Parse/extract our results
   -- Outcome {Success, Failure, Partial Success,
   N/A -- no results produced}

78 Phase 3 Step 3.1: Parse/extract our results -- Notes

79 Phase 3 Step 3.2: Compare to paper values
   -- Who did it {Human, Agent, Both,
   N/A -- no results produced}

80 Phase 3 Step 3.2: Compare to paper values
   -- Outcome {Success, Failure, Partial Success,
   N/A -- no results produced}

81 Phase 3 Step 3.2: Compare to paper values -- Notes

82 Phase 3 Step 3.3: Investigate discrepancies (if any)
   -- Who did it {Human, Agent, Both,
   N/A -- no results produced}

83 Phase 3 Step 3.3: Investigate discrepancies (if any)
   -- Outcome {Success, Failure, Partial Success,
   N/A -- no results produced}

84 Phase 3 Step 3.3: Investigate discrepancies (if any)
   -- Notes

85 Phase 3 Blocker 1: What was the blocker
86 Phase 3 Blocker 1: Who got stuck {Human, Agent, Both}
87 Phase 3 Blocker 1: What was the resolution
88 Phase 3 Blocker 1: Intervention needed? {Yes, No}

89 Phase 3 Blocker 2: What was the blocker
90 Phase 3 Blocker 2: Who got stuck {Human, Agent, Both}
91 Phase 3 Blocker 2: What was the resolution
92 Phase 3 Blocker 2: Intervention needed? {Yes, No}

93 Phase 3 Blocker 3: What was the blocker
94 Phase 3 Blocker 3: Who got stuck {Human, Agent, Both}
95 Phase 3 Blocker 3: What was the resolution
96 Phase 3 Blocker 3: Intervention needed? {Yes, No}

=========================================================
Part IV-A: Collaboration Patterns, Agent Contribution,
Struggle Analysis, and Reproduction Failure Classification
(Questions 97--101)
=========================================================

97 Collaboration pattern observed
   *If multiple choices apply, use the “Other” freeform text
   field.

   1. Agent did all the work on its own
   2. Agent asked for human input less than 5 times
   3. Human had to provide a minor suggestion or two to
      redirect agent on the right path
   4. Agent made major error(s), requiring human redirection
   5. Agent stopped before completing full answer(s),
      requiring human prodding to continue
   6. Agent asked for human input/assistance for several
      steps
   7. Agent and human worked back-and-forth as near-equal
      partners
   8. Agent completed task but required significant scope
      clarification upfront
   9. Agent failed completely
   10. Other:

98 Where Agent added value

   1. Navigating readme and necessary associated files
      quickly to understand requirements
   2. Environment setup
   3. Downloading data
   4. Identifying main scripts
   5. Running code
   6. Debugging errors from running code as is
   7. Making the most appropriate choice to adjust code to
      run correctly
   8. Identifying deprecated code/requirements and quickly
      finding fixes
   9. Catching potential issues proactively (e.g., noticing
      a bug in the code before it caused a major error)
   10. Finding an alternative more efficient approach
   11. Interpreting intermediate results intelligently so
       that it could move on quickly to next steps
   12. Other:

99 Where Agent struggled and needed help

   1. Understanding the initial prompt
   2. Following README instructions
   3. Setting up environment as directed in readme or
      repository
   4. Identifying data source and downloading it correctly
   5. Identifying correct scripts needed for reproducing
   6. Making appropriate adjustments for deprecated code
   7. Making an inappropriate adjustment to the source code
      for compatibility
   8. Providing the final answer
   9. Hallucinating file paths, function names or model
      details that didn't exist
   10. Losing track of context
   11. Not knowing when to stop and continuing past the
       correct solution
   12. Failure to produce final results, or to check 
   obviously incorrect intermediate results
   13. Getting stuck in a loop of retries
   14. Asking clarifying questions too late
   15. Making assumptions about the environment without
       checking
   16. Failure to follow instructions
   17. Other:

100 Reproduction failure mode classification
    (in case reproduction of the given target failed)

   1. Environment setup failure
   2. Missing dependencies
   3. Data access issues
   4. Ambiguous instructions
   5. Code bugs
   6. Conceptual misunderstanding
   7. Timeout / resource exhaustion
   8. Results do not match within error tolerance
   9. Other:

101 Other Notes (error messages, surprises, observations -
    anything that doesn't fit above)

=========================================================
Part IV-B: Reproduction Results and Execution Duration
(Questions 102--103)
=========================================================

102 Reproduction results

    Methodology

    Run once, round to same number of digits as paper’s
    value and check if it falls within floating point
    derived tolerance (e.g. 2.2e-11 = 0.000000000022 =
    1e5 * sys.float_info.epsilon). If yes, mark as Match

    If not, run twice more and use the CORE-Bench paper’s
    method to generate a tolerance interval from the three
    values (plus floating point derived tolerance). You can
    use this Colab notebook for calculating the interval.

    If the target value (from the paper) falls into this
    interval, mark as Within tolerance interval . If not, mark as Fail 

    Also refer to the instructions in the default agent
    prompt

    a) Results: Question
    b) Results: Paper Value
    c) Results: Our Value
    d) Results: Match {Match, Within tolerance interval, Fail}

103 Total duration (in minutes)

    Measure by: Start from difference between first and
    last timestamp (as provided by script), manually
    subtract lunch breaks etc. (afk time), add any
    additional time for analysis etc. after the last
    timestamp
\end{verbatim}

\begin{table}[t]
\label{appendix: process_success}
\caption{\textbf{Overview of reproduction outcomes by step.} Success indicates that the step was completed successfully, Partial Success indicates completion with runtime issues or interruptions, and Failed indicates unsuccessful completion. Agent refers to autonomous agent execution, Both refers to human--agent collaboration, and Human refers to human-only execution. N/A indicates that the step was not applicable (e.g., no discrepancy to investigate or no result available to assess).}
\smallskip
\centering
\footnotesize
\resizebox{0.99\linewidth}{!}{
\begin{tabular}{lccccccc}
\toprule
\textbf{Step} & \textbf{Agent\_Success} & \textbf{Agent\_Partial-Success} & \textbf{Both\_Success} & \textbf{Both\_Partial-Success} & \textbf{Both\_Failed} & \textbf{Human\_Success} & \textbf{N/A} \\
\midrule
1.1 Start the instance and Docker image$^{*}$ & 0 & 0 & 0 & 0 & 0 & 25 & 0 \\
1.2 Start Agent with logging and prompt replication target task$^{*}$ & 0 & 0 & 0 & 0 & 0 & 25 & 0 \\
1.3 Obtain the paper's code & 25 & 0 & 0 & 0 & 0 & 0 & 0 \\
1.4 Read README & 25 & 0 & 0 & 0 & 0 & 0 & 0 \\
1.5 Create environment & 25 & 0 & 0 & 0 & 0 & 0 & 0 \\
1.6 Install dependencies & 23 & 0 & 2 & 0 & 0 & 0 & 0 \\
1.7 Download/prepare data & 25 & 0 & 0 & 0 & 0 & 0 & 0 \\
1.8 Verify setup & 25 & 0 & 0 & 0 & 0 & 0 & 0 \\
2.1 Identify entry point / main script & 25 & 0 & 0 & 0 & 0 & 0 & 0 \\
2.2 Understand required run parameters & 24 & 0 & 1 & 0 & 0 & 0 & 0 \\
2.3 Run code & 20 & 2 & 3 & 0 & 0 & 0 & 0 \\
2.4 Monitor / debug runtime errors & 24 & 0 & 0 & 0 & 0 & 0 & 0 \\
2.5 Locate output files & 25 & 0 & 0 & 0 & 0 & 0 & 0 \\
3.1 Parse/extract our results & 24 & 0 & 0 & 0 & 0 & 0 & 1 \\
3.2 Compare to paper values & 22 & 0 & 2 & 0 & 0 & 0 & 1 \\
3.3 Investigate discrepancies & 17 & 1 & 3 & 0 & 0 & 0 & 4 \\
\bottomrule
\end{tabular}
}
\smallskip
\raggedright
\footnotesize{
$^{*}$These two steps were always executed by the human evaluator, by design.}

\label{tab:agent_added_value}
\end{table}

\begin{table}[t]
\centering
\caption{Evaluator-reported blockers in human-agent collaboration sessions}
\smallskip
\label{tab:blocker_summary}
\begin{tabular}{p{0.72\linewidth}r}
\toprule
Metric & Value \\
\midrule
Sessions with at least 1 substantive blocker & 11 (44\%) \\
Total substantive blocker events & 30 \\
Sessions with at least 1 blocker requiring human intervention & 5 (20\%) \\
Blocker events requiring human intervention & 10 (33\%) \\
Mean blocker events per affected session & 2.73 \\
\bottomrule
\end{tabular}

\vspace{0.5ex}
\begin{minipage}{0.92\linewidth}
\footnotesize
\textit{Note.} Blocker items were annotated only for human-agent collaborative sessions. Percentages are therefore calculated over human-agent collaborative sessions ($N=25$) or blocker events ($N=30$), as appropriate. One missing intervention flag was adjudicated as requiring intervention based on its description.
\end{minipage}
\label{tab:blocker_summary}
\end{table}

\begin{table*}[t]
\centering
\small
\setlength{\tabcolsep}{4pt}
\caption{Illustrative evaluator-reported blockers in human-agent collaborative sessions}
\label{tab:blocker_examples}
\begin{tabular}{@{}>{\raggedright\arraybackslash}p{0.36\textwidth}>{\raggedright\arraybackslash}p{0.38\textwidth}>{\centering\arraybackslash}p{0.12\textwidth}@{}}
\toprule
Example blocker & Resolution & Intervention? \\
\midrule
"The first run failed before the code started because this container doesn’t have /usr/bin/time." (according to the agent) & "I’m rerunning the same preprocessing command without that wrapper." & No \\
\midrule
According to the agent: "nltk==3.9 imports wordnet at module import time in this environment, so the original script stops before  preprocessing begins." & According to the agent: "I’ve hit the same nltk import bug twice now. One final environment-only fix is reasonable here: swap nltk to 3.8.1, which still provides the nltk.util.ngrams API this script uses but avoids the unrelated wordnet import-time failure on this Python 3.11 setup." & No \\
\midrule
Agent ran code on wrong dataset sample & Told agent to consult paper for dataset config. & Yes \\
\midrule
"package ‘oglmx’ is not available for this version of R" (and similar for others) & removed as unnecessary for replication target & No \\
\midrule
The code hit a difficult looking bug involving exhaustion of the C stack. & The agent stopped to check in with the user (as requested in the prompt), and suggested resorting to the older R version specified in the README, which worked after approval by the analyst & Yes \\
\midrule
The agent began with a smoke test and then paused to request guidance on next steps, likely recognizing that training all 40 models from scratch would be computationally intensive. & The agent estimated that completing the full training would take over 10 days, which exceeded available resources. Based on this constraint, the agent and the human researcher shifted the approach to using pretrained checkpoints to assess reproducibility. & Yes \\
\midrule
Agent misinterpreted the instructions and ran models with hyperparameters in the repo. & Human researcher advised the agent to follow the original instructions provided in the prompt: reproduce paper results & Yes \\
\bottomrule
\end{tabular}

\vspace{0.5ex}
\begin{minipage}{0.94\textwidth}
\footnotesize
\textit{Note.} Entries are reproduced verbatim from evaluator responses, except for LaTeX escaping and line wrapping. Examples were selected to illustrate the range of blockers and are not exhaustive.
\end{minipage}
\end{table*}

\begin{table}[t]
\caption{\textbf{Where the agent was perceived to be useful for human-agent collaborative reproduction runs.} Multiple selections were allowed per run. We consider an agent to be useful at a particular step in the human-agent collaboration runs based on the reproducer's judgement of steps they would have found difficult to fix without agent assistance.}
\smallskip
\centering
\footnotesize
\resizebox{0.95\linewidth}{!}{
\begin{tabular}{lc}
\toprule
\textbf{Where Agent added value} & \textbf{Mentions across runs} \\
\midrule
Environment setup & 25 \\
Running code & 23 \\
Identifying main scripts & 20 \\
Navigating readme and necessary associated files quickly to \\understand requirements & 19 \\
Downloading data & 17 \\
Debugging errors from running code as is & 14 \\
Making the most appropriate choice to adjust code correctly & 10 \\
Catching potential issues proactively (e.g., noticing a bug in the \\ code before it caused a major error) & 8 \\
Finding an alternative more efficient approach & 8 \\
Identifying deprecated code/requirements and quickly finding fixes & 7 \\
Interpreting intermediate results intelligently so that it could move on\\ quickly to next steps & 6 \\
\bottomrule
\end{tabular}
}
\label{tab:agent_added_value}
\end{table}

\begin{table}[t]
\caption{\textbf{Where the agent encountered difficulties across human-agent collaboration reproduction runs.} Multiple selections were allowed per run. Fourteen runs reported none of the following areas.}
\smallskip
\centering
\footnotesize
\resizebox{0.98\linewidth}{!}{
\begin{tabular}{lc}
\toprule
\textbf{Where the agent struggled} & \textbf{Mentions across runs} \\
\midrule
Identifying correct scripts needed for reproduction & 2 \\
Providing the final answer & 2 \\
Setting up environment as directed in the README/repository & 2 \\
Making assumptions about the environment without checking & 1 \\
Failure to follow instructions & 1 \\
Understanding the initial prompt & 1 \\
Making inappropriate compatibility adjustments to source code & 1 \\
Spending too much time pursuing an incorrect path & 1 \\
Forgetting original instructions and rescoping the task & 1 \\
Making a decision for the next step & 1 \\
Making appropriate adjustments for deprecated code & 1 \\
Losing track of context & 1 \\
Minor output formatting issues & 1 \\
\bottomrule
\end{tabular}
}
\label{tab:agent_struggles}
\end{table}

\begin{table}[t]
\caption{\textbf{Target-reproduction comparison between human-agent collaborative and manual reproduction runs.} Result category shows whether the reproduction attempt resulted in a final value for the reproduction target that matched with selected target from the published paper; either exactly or within a tolerance interval (see calculation in \ref{appendix:default_prompt}). If a manual run or human-agent collaborative run determined that the pipeline to reproduce the target value was not present in the code provided, the result was marked as a fail. Five manual runs were marked as failures solely because they exceeded the 3-hour runtime limit.} 
\label{appendix: match_comparison}
\smallskip
\centering
\footnotesize
\resizebox{0.75\linewidth}{!}{
\begin{tabular}{lcc}
\toprule
\textbf{Result category} & \textbf{Human-agent Collaborative} & \textbf{Manual} \\
\midrule
Exact match & 15 & 11 \\
Within tolerance interval & 3 & 4 \\
Fail & 7 & 10 \\
\bottomrule
\end{tabular}
}
\end{table}

\begin{table}[t]
\caption{\textbf{Additional evaluator observations from human-agent collaboration reproduction runs.} Most runs required no notable intervention and were completed successfully by the agent. Reported observations primarily related to execution efficiency, scope interpretation, runtime optimization, and the agent’s handling of discrepancies or recovery from initial errors. We reported no additional notes for 20 runs.}
\smallskip
\centering
\footnotesize
\resizebox{0.98\linewidth}{!}{
\begin{tabular}{p{0.97\linewidth}}
\toprule
\textbf{Other notes} \\
\midrule

Although the agent sought human guidance for the next step, it showed reasonable judgment by recognizing the computational cost and avoiding full pretraining, which would have required more than 10 days. \\
\midrule
Highly efficient agent run that successfully reproduced the result \\
\midrule
It was still somewhat impressive seeing the agent work its way through resolving the problems resulting from its initial wrong choice, and the eventual successful option went smoothly. Still, the [agent] could have saved over half an hour by following the README instructions (on the required R version etc.) more closely from the beginning. \\
\midrule
This particular reproduction went over the 45 minute compute time limit that was imposed as a criterion in the paper selection. I haven't investigated whether the agent could have chosen a more performant (e.g. multi-core) way to run the process. For the two additional runs required after the first result mismatch, it intelligently found a way to make them run in parallel so that they only required about the same time together as the first one alone.\\
\midrule
As described in more detail in the notes for [reproduction step] 3.3, on human request the agent was also helpful in investigating the discrepancy of the reproduced result and narrow[ing] down the possible cause. (Since this task is not explicitly specified in our prompt, I still rate this run as "Agent did all the work on its own".)\\
\midrule
a very smooth run by the agent\\

\bottomrule
\end{tabular}
}
\label{tab:other_notes_raw}
\end{table}

\subsection{Randomized study observations}
\label{appendix:human_agent_example}

We provide a few examples of instances where the agent overcame an operational blocker:

\begin{enumerate}
    \item In \textit{Beyond Accuracy Behavioral Testing of NLP Models with CheckList}, the run only progressed after rebuilding an older Python stack so the released suite could deserialize.
    \item In \textit{Yellow Vests, Pessimistic Beliefs, and Carbon Tax Aversion}, the agent had to move to an older \texttt{R 4.0.3} environment after the modern stack repeatedly failed.
    \item In \textit{Multiracial Identity and Political Preferences} and \textit{Informer}, the agent recreated expected filesystem layouts or traced historical code paths before the relevant pipeline could be evaluated.
\end{enumerate}

\subsection{Scaffold- and model-level failure mode decomposition examples by capsule}
\label{appendix:by_capsule}
\begin{figure}[t]
\centering
\includegraphics[width=0.7\linewidth]{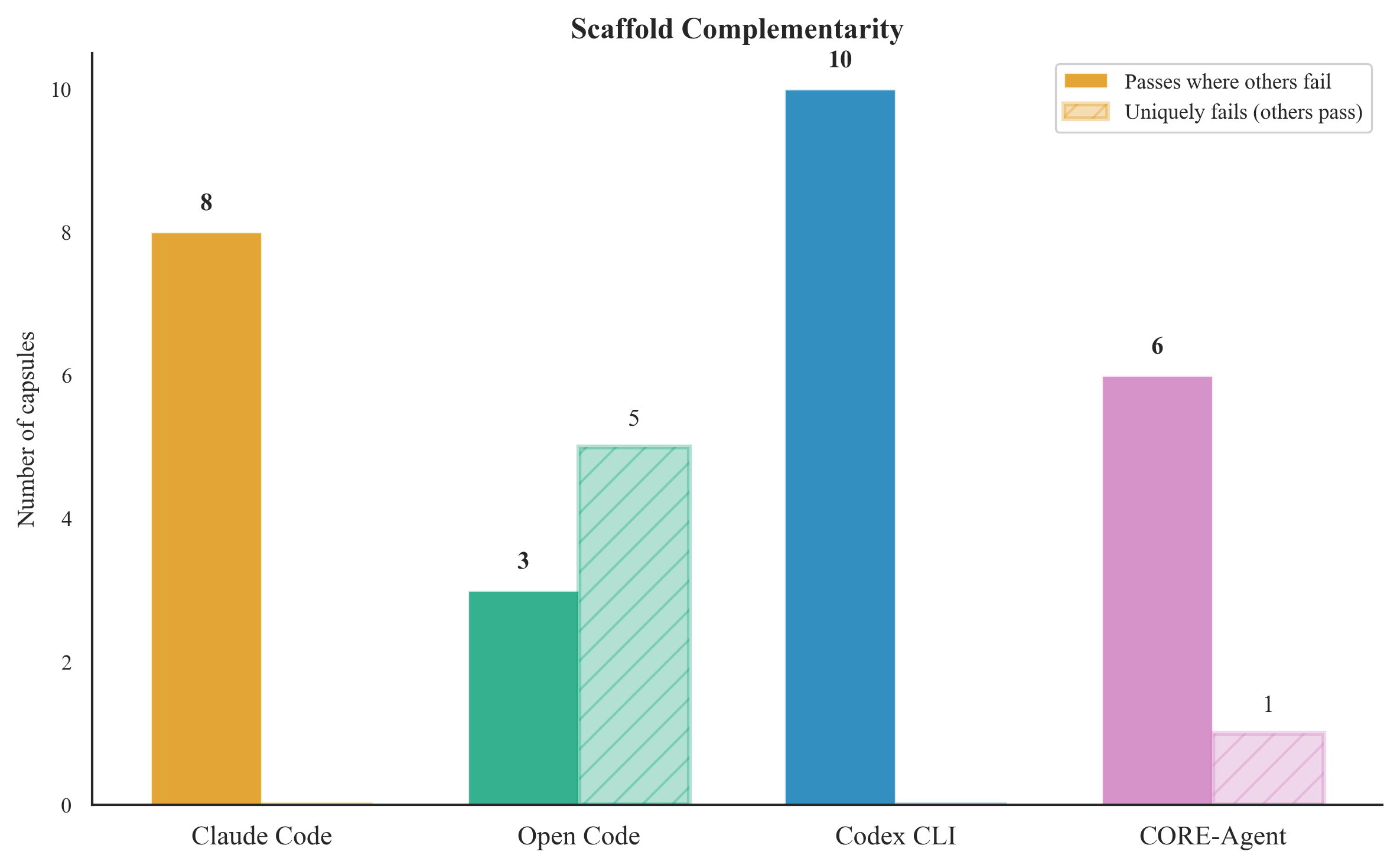}
\caption{\textbf{Scaffold complementarity across capsules.} Solid bars are cases where a scaffold passes while at least one other scaffold fails. Hatched bars are cases where the scaffold uniquely fails while others pass. Codex CLI provides the largest number of rescues with no unique failures in this slice, while CORE-Agent rescues some capsules but also uniquely fails others.}
\label{fig:scaffold_saves}
\end{figure}

\begin{figure}[t]
\centering
\includegraphics[width=\linewidth]{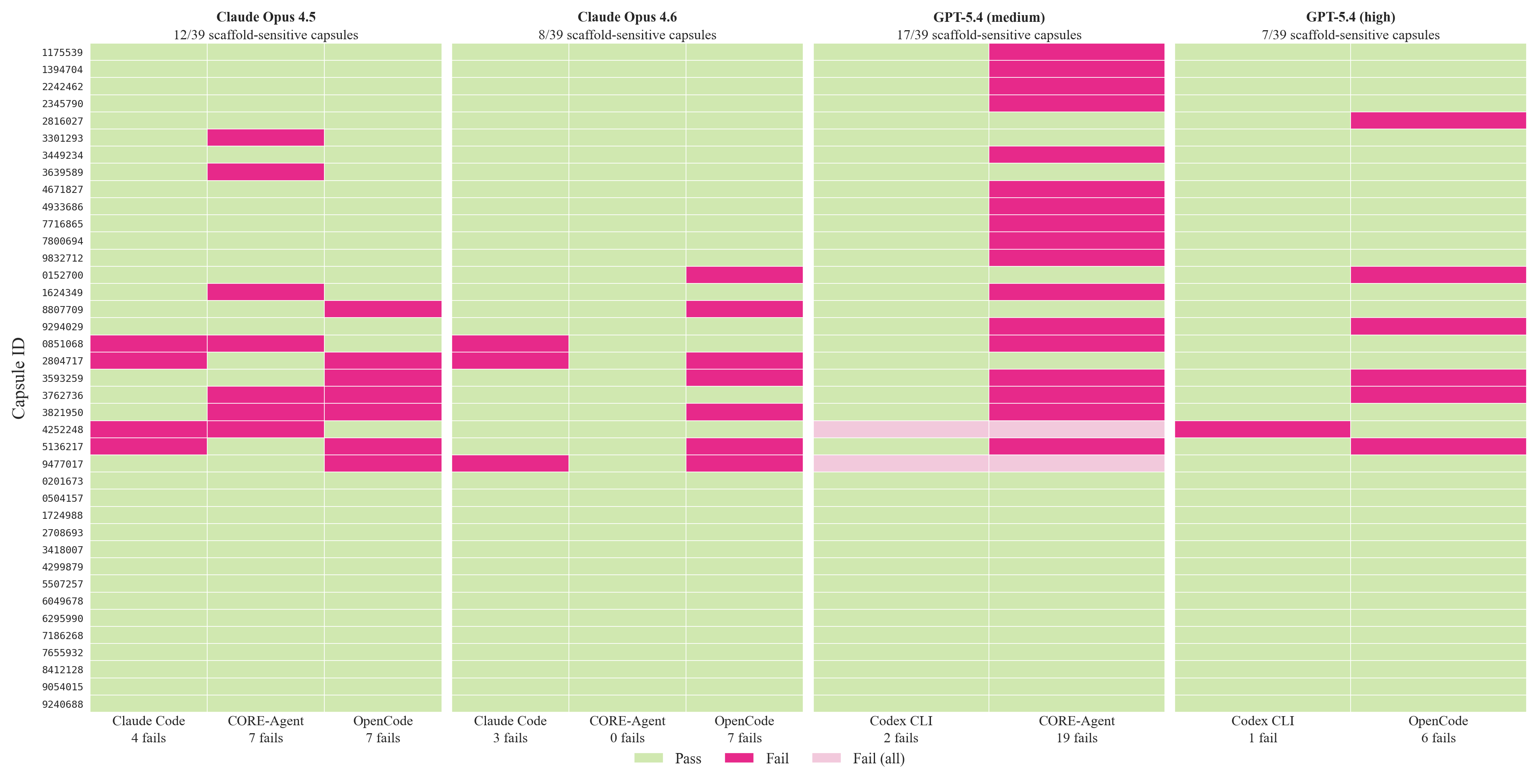}
\caption{\textbf{Per-capsule outcomes across scaffolds for the same model.} Each row is a capsule; each column is a scaffold. GPT-5.4 (medium) has the most scaffold-sensitive tasks (17/39), driven largely by CORE-Agent's 19 failures compared to Codex CLI's 2. Claude Opus 4.5 shows 12/39 scaffold-sensitive tasks, indicating that task-level disagreement can be substantial even when aggregate accuracy is similar.}
\label{fig:passfail_scaffold}
\end{figure}

\begin{figure}[t]
\centering
\includegraphics[width=\linewidth]{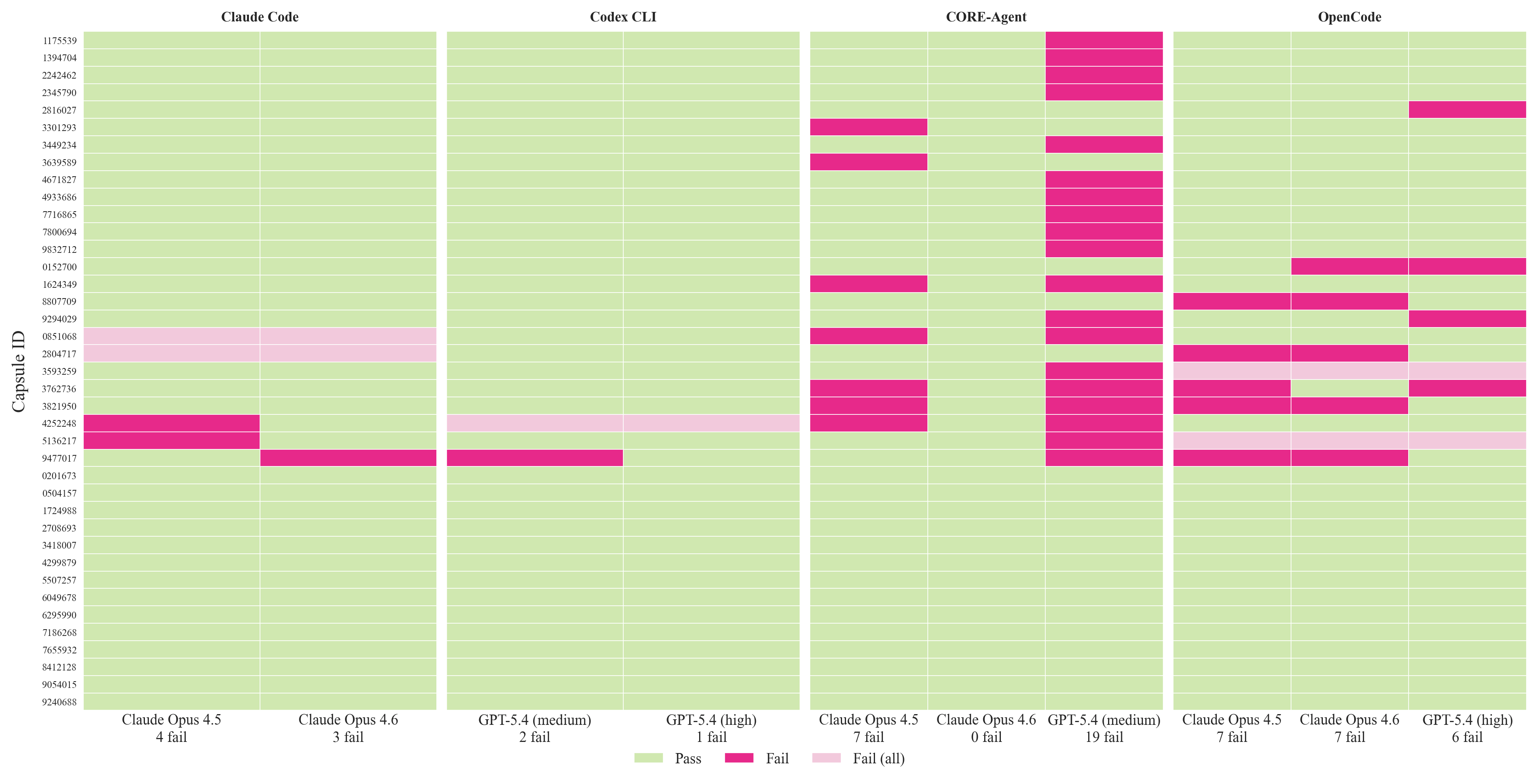}
\caption{\textbf{Per-capsule outcomes across models for the same scaffold.} Each row is a capsule; each column is a model. CORE-Agent shows the widest model sensitivity, with Claude Opus 4.6 passing all 39 tasks compared to 19 failures for GPT-5.4 (medium). Claude Code and Codex CLI show high model agreement, with near-identical failure patterns across their respective model pairs.}
\label{fig:passfail_model}
\end{figure}

We decompose accuracy along model and scaffold in \Cref{fig:scaffold_saves}, \label{sec:acc_decomp}, \Cref{fig:passfail_scaffold}, and \Cref{fig:passfail_model}. We present the following additional findings:

\textbf{Changing the scaffold alone can rescue performance.} In \texttt{capsule-1175539}, CORE-Agent's output format triggers early termination before the intended R analysis runs, while Codex CLI provides enough iteration budget for the same model to adapt to the library-path issue and complete the pipeline. This pattern extends broadly: 18 of GPT-5.4's 19 CORE-Agent failures recover in at least one Codex CLI configuration (17 at matched reasoning effort), with zero regressions. Message counts reinforce this reading: GPT-5.4 averages 36.8 messages on passing CORE-Agent runs and 36.0 on failing ones, suggesting the model does not change effort in response to difficulty. The recovered runs also require a message budget comparable to capsules that pass in both scaffolds (75 vs. 70), consistent with scaffold-imposed constraints rather than intrinsic task difficulty. Every one of GPT-5.4's 19 CORE-Agent failures passes under at least one alternative scaffold; none is a universal failure.

\textbf{Some trajectories follow the model, not the scaffold.} For \texttt{capsule-4252248}, Opus 4.6 computes the correct value (0.4929241) in three separate scaffolds, then submits the rounded figure-legend value (0.493) each time. GPT-5.4 and Opus 4.5, both with OpenCode, extract the value directly from code output without consulting the figure. The behavior recurs across scaffolds, pointing to a possible model-level tendency.

\textbf{Some failures depend on the match between agent speed and scaffold constraints.} In \texttt{capsule-5136217}, Claude Code with Opus 4.6 resolves the task in 63 messages and never encounters the \texttt{bsts}-dependent code, while Opus 4.5 in the same scaffold spends most of its 262 messages on package installation and is cut off by the 2,700s timeout before answer collection. In \texttt{capsule-0851068}, the pattern reverses: Claude Code with Opus 4.6  correctly diagnoses a PyTorch socket-path bug and computes the right AUC, but the timeout expires before the answer is submitted, while the same model in OpenCode reaches the fix faster and completes within budget. In both cases the model can solve the task; whether it finishes depends on how its working pace aligns with the scaffold's time limit.

\end{document}